\documentclass[lettersize,journal]{IEEEtran}
\usepackage{amsmath,amsfonts}
\usepackage{amssymb} % 提供额外的数学符号（可选）
\usepackage{algorithmic}
\usepackage{array}
\usepackage[caption=false,font=normalsize,labelfont=sf,textfont=sf]{subfig}
\usepackage{textcomp}
\usepackage{stfloats}
\usepackage{times} 
\usepackage{url}
\usepackage{verbatim}
\usepackage{graphicx}
\usepackage{cite}
\usepackage{color}
\usepackage{makecell}
\usepackage{bigstrut, multirow, rotating}
\usepackage{wrapfig}
\usepackage{booktabs}
\usepackage{bbm}
\usepackage{graphicx} % 用于插入图片
\usepackage{tcolorbox} % 用于创建彩色框
\tcbuselibrary{skins} % 提供更多样式选项
\definecolor{myred}{rgb}{1,0,0}
\usepackage[ruled,linesnumbered]{algorithm2e}

\newcommand{\ie}{\textit{i.e.}}
\newcommand{\eg}{\textit{e.g.}}
\definecolor{mygray}{gray}{.9}
\definecolor{myblue}{HTML}{1F77B4}
\definecolor{myyellow}{HTML}{FFA500}
\definecolor{lightblue}{HTML}{00ffff}
\definecolor{lightyellow}{HTML}{ffff00}
\begin{document}

\title{Background Matters: A Cross-view Bidirectional Modeling Framework for Semi-supervised \\Medical Image Segmentation}

\author{Luyang Cao, Jianwei Li,  Yinghuan Shi
        % <-this % stops a space
\thanks{This work is supported by NSFC Project (62222604, 62206052), China Postdoctoral Science Foundation (2024M750424), Fundamental Research Funds for the Central Universities (020214380120, 020214380128), State Key Laboratory Fund (ZZKT2024A14), Postdoctoral Fellowship Program of CPSF (GZC20240252), Jiangsu Funding Program for Excellent Postdoctoral Talent (2024ZB242) and Jiangsu Science and Technology Major Project (BG2024031).}
\thanks{Luyang Cao is with the State Key Laboratory for Novel Software Technology, Nanjing University, Nanjing, Jiangsu, China; the College of Physics and Information Engineering, Fuzhou University, Fuzhou, Fujian, China; and the National Institute of Healthcare Data Science, Nanjing University, Nanjing, Jiangsu, China; (e-mail: caoluyang@smail.nju.edu.cn).}

\thanks{Jianwei Li is with the College of Physics and Information Engineering, Fuzhou University, Fuzhou 350108, China (e-mail: lwticq@163.com).}
\thanks{Yinghuan Shi is with the National Key Laboratory for Novel Software Technology, National Institute of Healthcare Data Science, Nanjing University, and Nanjing Drum Tower Hospital, Nanjing, Jiangsu, China (e-mail: syh@nju.edu.cn).}
\thanks{*The corresponding author: Jianwei Li, Yinghuan Shi.}}

\markboth{ }%
{Shell \MakeLowercase{\textit{et al.}}: A Sample Article utilizing IEEEtran.cls for IEEE Journals}

\IEEEpubid{0000--0000/00\$00.00~\copyright~2024 IEEE}
% Remember, if you utilize this you must call \IEEEpubidadjcol in the second

\maketitle

\begin{abstract}
Semi-supervised medical image segmentation (SSMIS) leverages unlabeled data to reduce reliance on manually annotated images. However, current SOTA approaches predominantly focus on foreground-oriented modeling (\ie, segmenting only the foreground region) and have largely overlooked the potential benefits of explicitly modeling the background region.
{Our study theoretically and empirically demonstrates that highly certain predictions in background modeling enhance the confidence of corresponding foreground modeling.} Building on this insight, we propose the Cross-view Bidirectional Modeling (CVBM) framework, which introduces a novel perspective by incorporating background modeling to improve foreground modeling performance. Within CVBM, background modeling serves as an auxiliary perspective, providing complementary supervisory signals to enhance the confidence of the foreground model. Additionally, CVBM introduces an innovative bidirectional consistency mechanism, which ensures mutual alignment between foreground predictions and background-guided predictions.
Extensive experiments demonstrate that our approach achieves SOTA performance on the LA, Pancreas, ACDC, and {HRF datasets}. Notably, on the Pancreas dataset, CVBM outperforms fully supervised methods (\ie, DSC: 84.57\% vs. 83.89\%) while utilizing only 20\% of the labeled data. Our code is publicly available at \color{magenta}{https://github.com/caoluyang0830/CVBM.git}.
\end{abstract}


\begin{figure}[t]
\centering

\includegraphics[width=1.0\columnwidth]{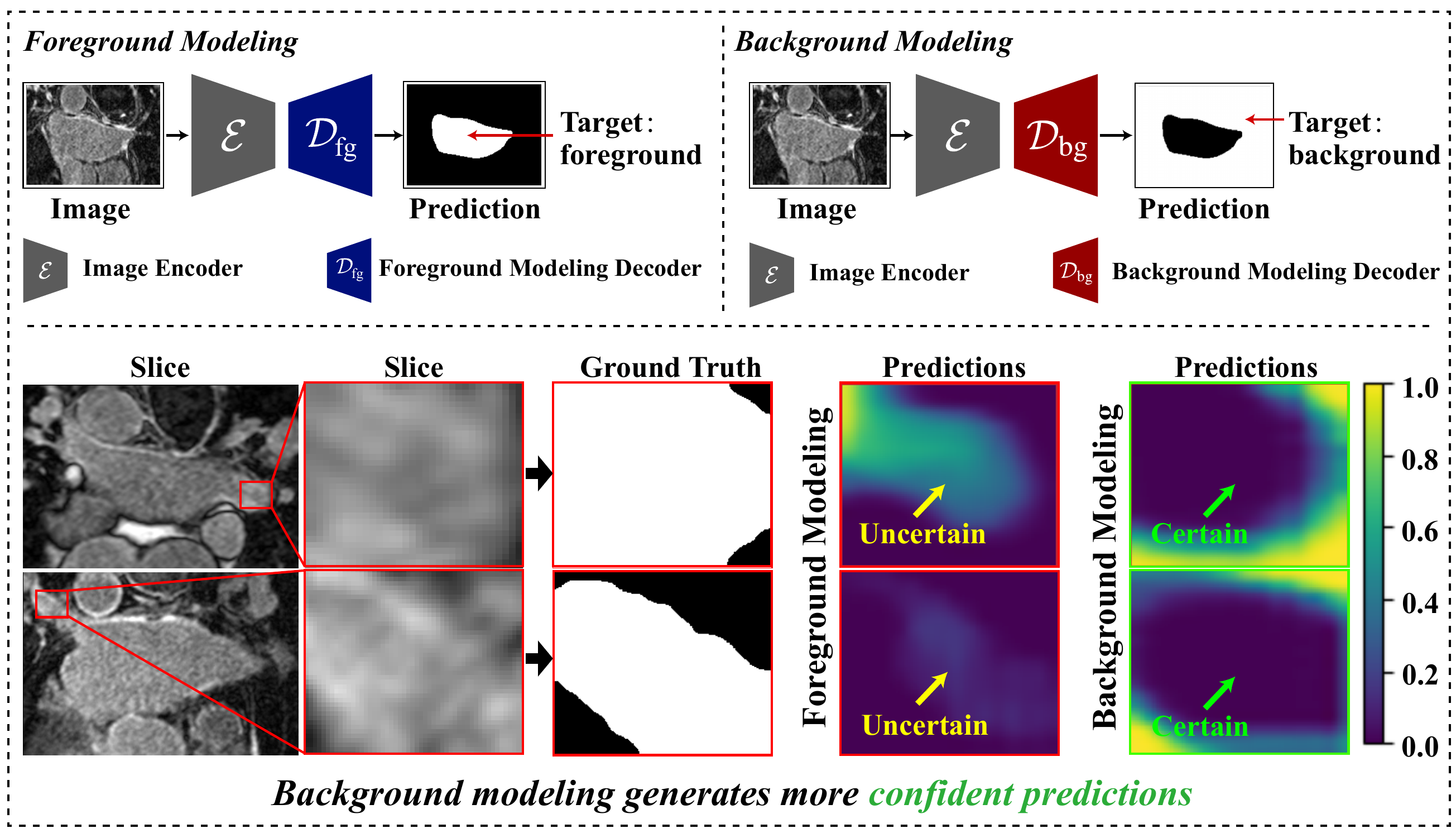} % Reduce the figure size so that it is slightly narrower than the column. Don't utilize precise values for figure width.This setup will avoid overfull boxes.
\caption{The motivation of proposed approach. {{In some cases}, background modeling exhibits higher predictive confidence compared to foreground modeling. The upper panel illustrates the conceptual definitions of foreground and background modeling, while the lower panel depicts the predictions from each modeling scheme. Both foreground and background models were trained utilizing VNet~\cite{milletariVNetFullyConvolutional2016} on the LA dataset.}}
\label{fig:FBVNet}

\end{figure}

\section{Introduction} \label{sec:Introduction}

\label{sec:intro}
\IEEEPARstart{M}{ainstream} deep learning-based segmentation models have demonstrated {effectiveness in achieving} precise segmentation. {However, the success of these models heavily relies on the availability of} pixel-level annotations~\cite{liuGroupingBoundaryProposals2024,sunMASCLEndtoendMultiatlas2024,wuRobustReferringImage2024}. {Unfortunately, acquiring dense} pixel-level labels is {highly costly and labor-intensive, particularly in medical imaging}~\cite{liuBilateralContextModeling2024,liuMMNetMixformerbasedMultiscale2024,luAnomalyDetectionMedical2024,sunMASCLEndtoendMultiatlas2024a,WANGyou_2025_TMI,WANG_2025_TMI,li_2025_TMI,Ma_2024_CVPR}. {Semi-Supervised Medical Image Segmentation (SSMIS) offers a promising solution by leveraging} limited labeled data and abundant unlabeled data, {significantly reducing annotation costs}~\cite{liuSemiRSCOCSemiSupervisedClassification2024,wangHybridPerturbationStrategy2024,weiSemiSupervisedLearningHeterogeneous2024}. In SSMIS, numerous methods have been {developed to effectively segment organs and lesion areas (referred to as foreground regions). These methods are broadly categorized into two paradigms:} entropy minimization and consistency regularization. Entropy minimization techniques encourage models to produce high-confidence (low-entropy) predictions for foreground regions~\cite{liuACPLAnticurriculumPseudolabelling2022,zhangMultimodalContrastiveMutual2023}. {Consistency regularization methods minimize discrepancies between multiple predictions of the same sample under different perturbations, generating more reliable segmentation results}~\cite{wangSemisupervisedMedicalImage2022,wuMutualConsistencyLearning2022,liuPerturbedStrictMean2022}. {Furthermore, recent advancements integrating both paradigms have demonstrated significant performance improvements, highlighting the potential of hybrid approaches}~\cite{chenMagicNetSemiSupervisedMultiOrgan,wuCrosspatchDenseContrastive2022}.

\IEEEpubidadjcol
Despite their success, we have observed an important yet easily underestimated issue: {segmentation models predominantly focus on foreground modeling (as illustrated in the upper left of Fig.~\ref{fig:FBVNet}) while neglecting explicit modeling of the background region (shown in the upper right of Fig.~\ref{fig:FBVNet}).} Moreover, during training, the background is frequently treated as a disturbance factor~\cite{pengSelfPacedContrastiveLearning2021,zhouSSMDSemiSupervisedMedical2021}, {seemingly irrelevant to foreground segmentation}. However, {accurate background segmentation inherently implies accurate foreground segmentation}. Naturally, a critical question arises: \textit{Is the background truly irrelevant? What role does it play in the modeling process?}

% \begin{figure}[t]
% \centering
% \includegraphics[width=1.0\columnwidth]{different framework.pdf} % 
% \caption{Our Cross-view Bidirectional Modeling scheme (CVBM). The symbol $\varepsilon$ signifies the encoder, ${{D}_\text{fg}}$ and ${{D}_\text{bg}}$ denote decoders for foreground and background feature learning, respectively. {(a) illustrates the definitions of foreground and background modeling, while (b) shows the main framework of CVBM.}}
% \label{fig:different_framework}
% \end{figure}

In our investigation, we uncover an intriguing phenomenon: {in certain cases,} background modeling exhibits higher predictive confidence {than} foreground modeling. We illustrate this phenomenon in Fig.~\ref{fig:FBVNet}. {The bottom image compares the prediction confidence between foreground and background modeling for the same slice. Notably, in peripheral regions, foreground voxels exhibit high similarity to background voxels. This inherent voxel similarity limits the segmentation model's ability to accurately identify foreground regions, resulting in uncertain predictions (highlighted by yellow arrows). For most SSMIS methods, generating foreground pseudo-labels typically involves filtering out uncertain predictions~\cite{phamMetaPseudoLabels2021,heSECRETSelfConsistentPseudo2022,nassarProtoconPseudoLabelRefinement2023}, which introduces discrepancies from the ground truth and reduces pseudo-label reliability. In contrast, when applying background modeling to the same image slices, regions previously identified as uncertain demonstrate higher predictive certainty (indicated by green arrows). Therefore, compared to foreground modeling, background modeling successfully identifies the foreground in these challenging regions. This suggests that background modeling enhance the predictive confidence of the foreground modeling.}
{In our theoretical analysis, we establish that highly certain predictions in background modeling enhance the confidence of corresponding foreground modeling (Appendix B). This phenomenon persists even in fully supervised segmentation. However, the extensive availability of labeled data in such methods diminishes the impact of uncertain predictions in foreground modeling. Building on this insight, we propose that in SSMIS, background modeling could be strategically leveraged to support foreground modeling, thereby reducing uncertain regions and improving the overall reliability of foreground segmentation.}

{In this work, we diverge from the prevailing trend in recent SOTA methods, which predominantly focus on foreground segmentation, by proposing a Cross-view Bidirectional Modeling (CVBM) framework. This approach integrates background modeling to enhance foreground segmentation performance. Specifically, CVBM concurrently analyzes input data from both foreground and background perspectives. While foreground modeling remains the primary objective for segmenting the target region, background modeling provides a complementary viewpoint to improve the predictive confidence of the foreground model. Additionally, we introduce a mixing layer to seamlessly integrate predictions from both models. To optimize the framework, we propose a bidirectional consistency mechanism, which enforces direct and inverse consistency constraints on foreground predictions. This mechanism reduces low-confidence regions, thereby enhancing segmentation reliability.}
To the best of our knowledge, this study represents the first investigation into the role of background modeling in SSMIS. Our primary contributions are summarized as follows:
\begin{itemize}
    \item {\textbf{Theoretical insight}: Highly certain predictions in background modeling are established to enhance the confidence of corresponding foreground modeling.}
    \item {\textbf{Novel modeling perspective}: A Cross-view Bidirectional Modeling (CVBM) framework is designed to leverage background modeling, assisting the foreground model in reducing uncertain regions.}
    \item {\textbf{Innovative consistency strategy}: A bidirectional consistency optimization mechanism is proposed to ensure mutual alignment between foreground predictions and background-guided predictions.}
\end{itemize}

{We evaluated the proposed method on four benchmarks in SSMIS: the LA, Pancreas, ACDC, and HRF datasets. Extensive experimental results demonstrate that CVBM outperforms SOTA algorithms. Notably, our method achieves superior performance even compared to fully supervised models on the Pancreas dataset, achieving a Dice Similarity Coefficient (DSC) of 84.57\% with only 20\% labeled data, surpassing the fully supervised baseline (DSC: 83.89\%).}

\section{Related Works}

\textbf{Semi-supervised Learning:} 
Semi-supervised learning (SSL) {encompasses diverse methodologies that leverage} abundant unlabeled data to improve model performance. A widely adopted SSL strategy is entropy minimization~\cite{YasarlaGaussian2021, yuCMOSGANSemiSupervisedGenerative2023, rizveDefensePseudoLabelingUncertaintyAware2021}, which {encourages networks to produce low-entropy predictions. This objective is achieved through explicit constraints~\cite{berthelotMixMatchHolisticApproach2019, wangAdversarialDenseContrastive2023, wuCrosspatchDenseContrastive2022} or by selecting high-confidence pseudo-labels for the foreground~\cite{rizveDefensePseudoLabelingUncertaintyAware2021, zhangOTAMatchOptimalTransport2024a, phamMetaPseudoLabels2021, kangDistillingSelfSupervisedVision2023}.} {The diversity of pseudo-labeling strategies has spurred the development of various frameworks, including multi-target optimization~\cite{panLearningSelfSupervisedLowRank2022}, geometry-aware methods~\cite{wangSemisupervisedMedicalImage2022}, and target edge detection~\cite{chenMultiTaskMeanTeacher2020}.}
Consistency regularization represents another cornerstone of SSL~\cite{kimSemiSupervisedLearningSemantic2022, geLearningPrivacyPreservingStudent2023, mittalSemiSupervisedSemanticSegmentation2021}, where models are trained to maintain consistent predictions under diverse perturbations. This principle has led to perturbation-based methodologies, \eg, data augmentation techniques~\cite{yiLearningPixelLevelLabel2022, zhangSamplecentricFeatureGeneration2022a, nassarProtoconPseudoLabelRefinement2023} and model perturbation strategies~\cite{wangConflictBasedCrossViewConsistency2023, chenSSLImprovingSelfSupervised2022a, wuMutualConsistencyLearning2022}. These SSL techniques provide a robust foundation for advancing semi-supervised medical image segmentation research.

\textbf{Semi-supervised Medical Image Segmentation: } 
{Building on SSL, numerous approaches have been proposed for SSMIS.} For perturbation-based modeling~\cite{liuSemiSupervisedMedicalImage2020, huangSemiSupervisedNeuronSegmentation2022, chenGenerativeConsistencySemisupervised2022}, {Luo \textit{et al.}\cite{luoEfficientSemisupervisedGross2021} {extend backbone segmentation networks to generate pyramid predictions at multiple scales. Similarly, Li \textit{et al.}~\cite{liDualConsistencySemisupervisedLearning2021} enforce both image transformation equivalence and feature perturbation invariance to effectively utilize unlabeled data.} In multi-task modeling~\cite{chaitanyaSemisupervisedTaskdrivenData2021, zhaoSemiSupervisedMedicalImage2023, zhangBoostMISBoostingMedical2022}, {Luo \textit{et al.}~\cite{luoSemisupervisedMedicalImage2021} propose a dual-task network that jointly predicts segmentation maps and geometry-aware representations of the foreground. Wang \textit{et al.}~\cite{wangSemisupervisedMedicalImage2022} integrate segmentation, reconstruction, and geometry-aware prediction tasks to refine foreground pseudo-labels.} {Additionally, co-training-based methods~\cite{liMultiviewCotrainingNetwork2023, liTransformationConsistentSelfEnsemblingModel2021, peirisUncertaintyguidedDualviewsSemisupervised2023} demonstrate effectiveness by leveraging multiple foreground models.}
{Despite their success, existing approaches primarily focus on foreground modeling, neglecting explicit background modeling. To address this gap, background modeling is incorporated as an auxiliary perspective, enabling foreground models to resolve ambiguous predictions autonomously. To the best of our knowledge, this work represents the first attempt to integrate background modeling into SSMIS.}

\textbf{Complementary Label: }\label{Complementary}
{Complementary labeling introduces an innovative paradigm in image classification, demonstrating cost-effectiveness and accuracy~\cite{duanMutexMatchSemiSupervisedLearning2022, rizveDefensePseudoLabelingUncertaintyAware2021, wangSemiSupervisedSemanticSegmentation2022}.} However, its exploration remains limited, with only a few SSL studies incorporating this paradigm. Different from conventional methods relying on original image labels (\ie, the category to which the image belongs), {complementary labeling directs models to focus on non-target classes (\ie, categories not belonging to the image).}
For instance, Duan \textit{et al.}~\cite{duanMutexMatchSemiSupervisedLearning2022} assign complementary labels to the class with the lowest predicted score, achieving 92.23\% accuracy on CIFAR-10 with only 20 labeled samples. Similarly, Rizve \textit{et al.}~\cite{rizveDefensePseudoLabelingUncertaintyAware2021} employ a threshold to filter predicted scores of multiple classes, integrating filtered classes as complementary labels, reducing the error rate by 10.86\% on CIFAR-10.
In image segmentation, complementary labeling remains under-explored, with limited relevant studies. Notably, Wang \textit{et al.}~\cite{wangSemiSupervisedSemanticSegmentation2022} investigate its potential in semi-supervised segmentation, utilizing adaptive thresholding to identify and designate unreliable voxels as complementary labels. Their approach outperforms the supervised baseline on the PASCAL VOC 2012 dataset under 1/16 partition protocols. These studies highlight the efficacy of complementary labeling in segmentation. However, its application in SSMIS remains unexplored, motivating our investigation. To the best of our knowledge, this work represents the first attempt to apply complementary labeling to SSMIS.

{\textbf{Background Modeling: }\label{Complementary}
In weakly supervised segmentation (WSS), effective background modeling is crucial for generating high-quality pseudo-masks from image-level labels. Recent studies have addressed the challenge of distinguishing foreground objects from co-occurring background pixels through various approaches. Yin \textit{et al.}\cite{yinFineGrainedBackgroundRepresentation2024} introduced negative regions of interest to contrast with confounding background pixels, while Zhai \textit{et al.}\cite{zhaiBackgroundActivationSuppression2024} proposed an activation map constraint module to suppress background activation values. Chen \textit{et al.}\cite{chenSpatialStructureConstraints2024} employed spatial structure constraints through a CAM-driven reconstruction module to preserve image spatial structure and prevent object over-activation into background regions. Contrastive learning techniques have also proven effective, with pixel-to-prototype contrast methods\cite{duWeaklySupervisedSemantic2022} improving feature space organization, and Xie \textit{et al.}\cite{xieC2AMContrastive2022} leveraging unlabeled data to disentangle foreground from background based on semantic differences. Foundation models have further advanced this area, with Yang \textit{et al.}\cite{yangFoundationModelAssisted2024} combining CLIP and SAM to progressively refine background representation. However, existing methods primarily focus on foreground-background contrast or background feature suppression. Our approach differs by identifying the inherent bidirectional consistency between foreground and background segmentation tasks and explicitly modeling background regions. We theoretically demonstrate that background modeling enhances foreground prediction confidence.}
\section{Method}

{In this section, {we first provide an overview of our method in Section~\ref{sub:Method Overview}.} We then present the proposed Cross-view Bidirectional Modeling (CVBM) framework, which consists of three key components: 1) background label settings (Section~\ref{sub:Complementary Label}), 2) the cross-view bidirectional modeling architecture (Section~\ref{sub:network}), and 3) the bidirectional optimization strategy (Section~\ref{sub:bidirectional consistency}).}

{\subsection{Method Overview} \label{sub:Method Overview}
As illustrated in Fig.~\ref{fig:overview}, we propose a cross-view bidirectional model (CVBM) that simultaneously models both foreground and background regions. To fully leverage bidirectional modeling capabilities in the SSMIS domain, we implement CVBM in both teacher and student roles. Our method follows a two-phase approach. Initially, we pre-train the teacher CVBM utilizing foreground and background labels on annotated data. Through iterative training, the teacher model tends to produce labels that are close to the ground truth. Subsequently, we utilize this pre-trained teacher model to generate foreground and background pseudo-labels across both labeled and unlabeled datasets, which are then utilized to train the CVBM student model via bidirectional consistency optimization. In the following sections, we detail three key components of our method: settings of background label, cross-view bidirectional modeling, and bidirectional consistency optimization.}
\begin{figure}[t]
    \centering
        \includegraphics[width=0.95\linewidth]{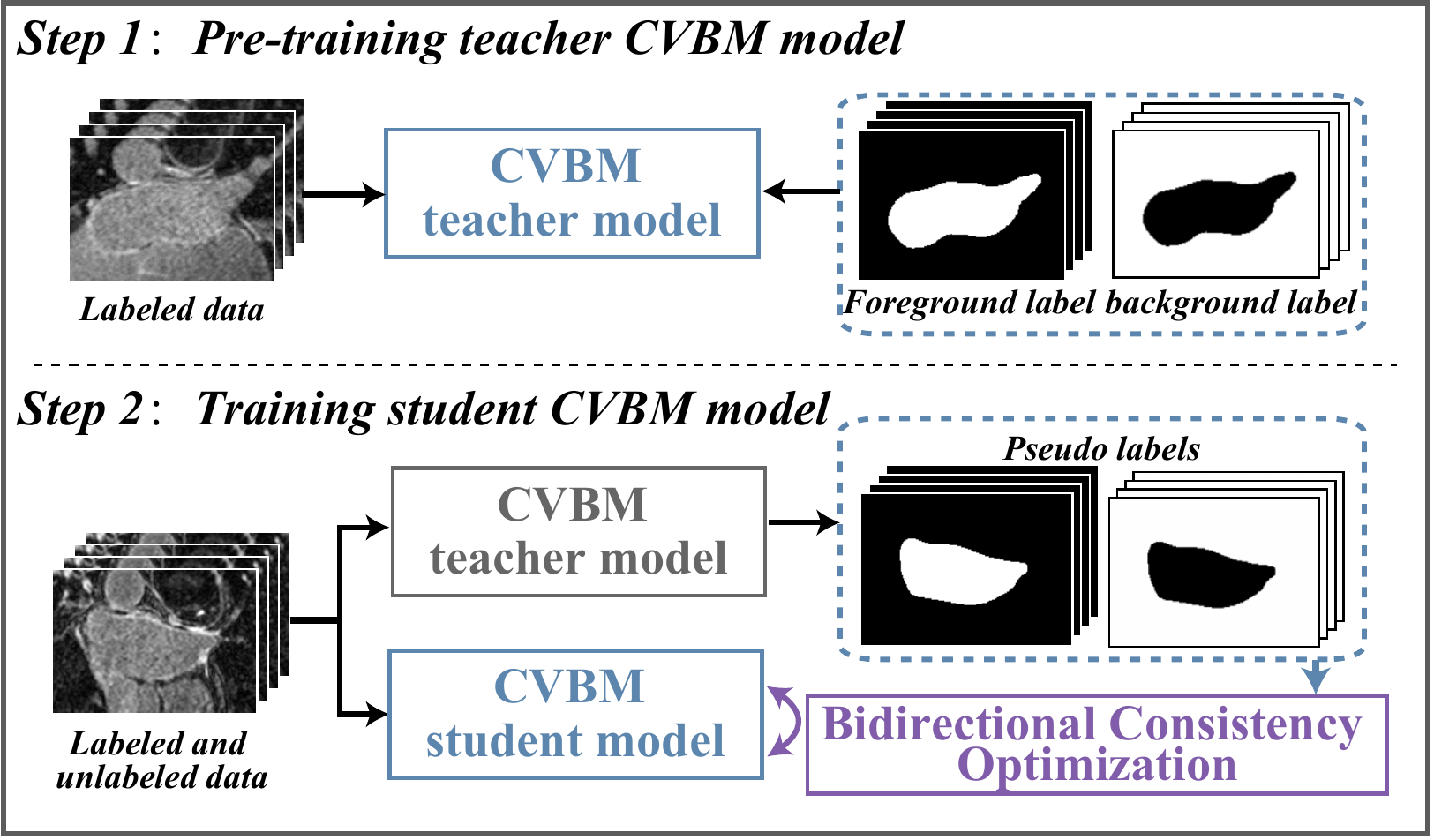} % \linewidth 自动适应 tcolorbox 宽度

    \caption{Overview of our proposed method. Model in gray represent stop gradient operations.}

    \label{fig:overview}
\end{figure}
\subsection{Settings of Background Label} \label{sub:Complementary Label}
{Complementary labels have proven effective in classification and segmentation tasks for 2D natural images. However, their direct application to 3D medical image segmentation faces challenges such as three-dimensional mismatch, modality diversity, and class similarity. This raises the question of whether a standardized definition of complementary labels can be established for 3D medical image segmentation. To address this issue, as illustrated in Fig.~\ref{fig:BGlabel}, we propose a definition of background labels specifically tailored for segmenting background regions in medical images. We term these labels as auxiliary complementary labels, which are applicable to both single-target and multi-target segmentation scenarios.}
\begin{figure}[t]

\centering
\includegraphics[width=0.95\columnwidth]{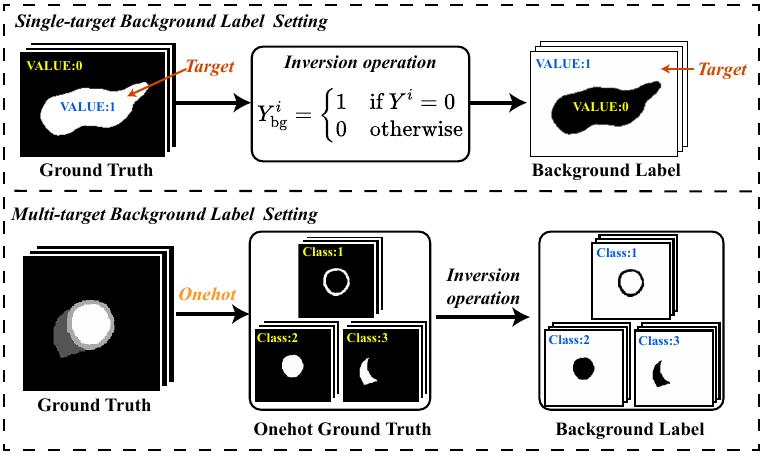}

\caption{{Background label settings. The inversion operation transforms binary label representations, converting background label values from 0 to 1 and foreground label values from 1 to 0. For single-target background labels, this operation is applied directly. For multi-target background labels, one-hot encoding is performed prior to inversion.}}

\label{fig:BGlabel}
\end{figure}
\subsubsection{{Single-Target Segmentation}}
{The ground truth is represented as a binary tensor ${Y} \in \{0, 1\}^{w \times h \times d}$, where $w$, $h$, and $d$ denote the width, height, and depth of the volume, respectively. Foreground voxels are set to 1, while background voxels are set to 0. Auxiliary complementary labels are derived through binary inversion, as illustrated in the upper panel of Fig.~\ref{fig:BGlabel}. This operation is formally expressed as:}
\begin{equation}
{Y}_\text{bg}^{i} = \begin{cases}
1 & \text{if } {Y}^{i} = 0 \\[-0.3em] % 调整行间距
0 & \text{otherwise}
\end{cases},
\label{con:bg1}
\end{equation}
{where ${Y}_\text{bg}$ denotes the background label, and $i$ represents the voxel index. Specifically, ${Y}_\text{bg}^{i}$ is set to 1 when ${Y}^{i} = 0$.}

\subsubsection{{Multi-Target Segmentation}}
{The ground truth for multi-class segmentation is represented as ${Y}_\text{M} \in {0, 1, \ldots, K}^{w \times h \times d}$, where $K+1$ is the total number of categories. Since the ground truth contains multiple categories, direct inversion of the ground truth is inapplicable. Instead, we utilize one-hot encoding to convert each category into a binary representation, setting the index of the specific category to 1 and all others to 0. Auxiliary complementary labels are created by inverting these one-hot encoded labels. This process is formally expressed as:}
\begin{equation}
{Y}_\text{M,bg}^{i} = \begin{cases}
1 & \text{if } \text{onehot}({Y}_\text{M}^{i}) = 0 \\[-0.3em]
0 & \text{otherwise}
\end{cases},
\label{con:bg2}
\end{equation}
{where $\text{onehot}(\cdot)$ denotes the one-hot conversion operation. The resulting label ${Y}_\text{M,bg} \in \{0, 1\}^{w \times h \times d \times c}$ is a high-dimensional tensor, where $c$ represents the number of categories.}

\begin{figure*}[t]
\centering

\includegraphics[width=0.95\textwidth]{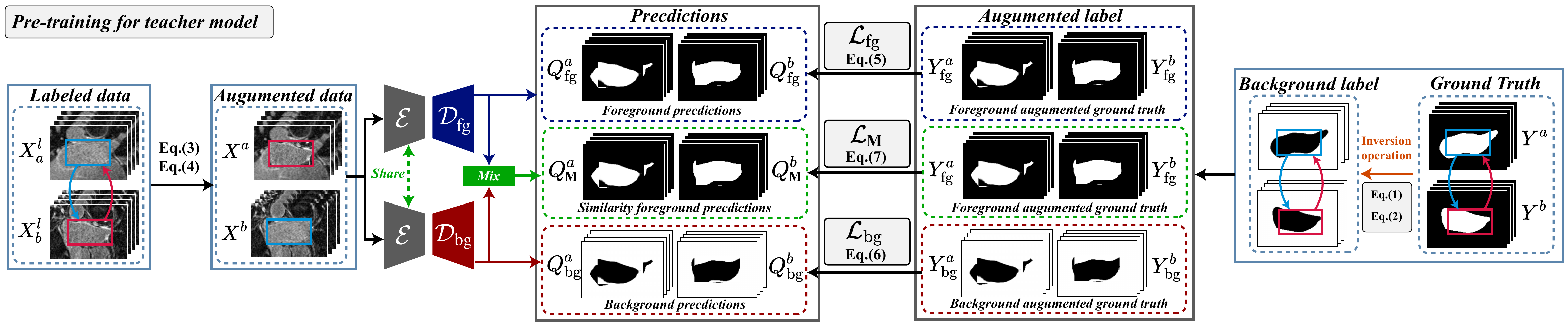} % Reduce the figure size so that it is slightly narrower than the column. Don't utilize precise values for figure width.This setup will avoid overfull boxes.

\caption{{Pre-training process of teacher model. For the training process of our teacher network, only labeled data are utilized for pre-training. The network processes cutmix inputs (${{{{X}}}^{a}}$, ${{{{X}}}^{b}}$)  and performs three core tasks: foreground modeling, background modeling and mixing, generating the respective predictions $Q_\text{fg}^a$ and $Q_\text{fg}^b$, $Q_\text{bg}^a$ and  $Q_\text{bg}^b$, $Q_\text{M}^a$ and $Q_\text{M}^b$. Our optimization involves minimizing the foreground segmentation loss ($\mathcal{L}_\text{fg}$), the background segmentation loss ($\mathcal{L}_\text{bg}$) and the mixed {prediction}  loss  ($\mathcal{L}_\text{M}$).} }

\label{fig:teacher}
\end{figure*}

\subsection{Cross-view Bidirectional Modeling}\label{sub:network}
To elucidate the intrinsic mechanisms of CVBM, we present a comprehensive analysis of the training processes for both the teacher and student models.
\subsubsection{Training Process of Teacher Model}
{Different from existing foreground modeling methodologies~\cite{wuCrosspatchDenseContrastive2022,wangSemisupervisedMedicalImage2022}, which primarily generate foreground pseudo-labels for unlabeled volumes, our approach introduces a bidirectional prediction mechanism. Specifically, the teacher model produces both foreground pseudo-labels $P_\text{fg}$ and background pseudo-labels $P_\text{bg}$ for each unlabeled volume. However, during initial training stages, the teacher network generates pseudo-labels with a high proportion of low-confidence voxels, insufficient to guide the student model effectively. To address this limitation, we pre-train the teacher model utilizing labeled data (\ie, ${X}^{l}_{a}$ and ${X}^{l}_{b}$) through our cross-view bidirectional modeling scheme, as illustrated in Fig.~\ref{fig:teacher}.}
{Cut-mix data augmentation~\cite{baiBidirectionalCopyPasteSemiSupervised2023} initializes the training data for the teacher model:}
\begin{equation}
{{X}^{a}} = {X}^{l}_{a} \odot \mathcal{M} + {X}^{l}_{b} \odot (1 - \mathcal{M}),
\label{con:bcp1}
\end{equation}
\begin{equation}
{{X}^{b}} = {X}^{l}_{a} \odot (1 - \mathcal{M}) + {X}^{l}_{b} \odot \mathcal{M},
\label{con:bcp2}
\end{equation}
{where ${{X}^{a}}$ and ${{X}^{b}}$ represent augmented labeled data, $\odot$ denotes element-wise multiplication, and $\mathcal{M} \in \{0, 1\}^{w \times h \times d}$ is the binary mask used to cut sub-volumes. The size of the zero-valued region in $\mathcal{M}$ is $\beta w \times \beta h \times \beta d$, with $\beta$ set to $2/3$~\cite{baiBidirectionalCopyPasteSemiSupervised2023}. This process is illustrated in Fig.~\ref{fig:cutmix}.}

\begin{figure}[t]
\centering

\includegraphics[width=0.93\columnwidth]{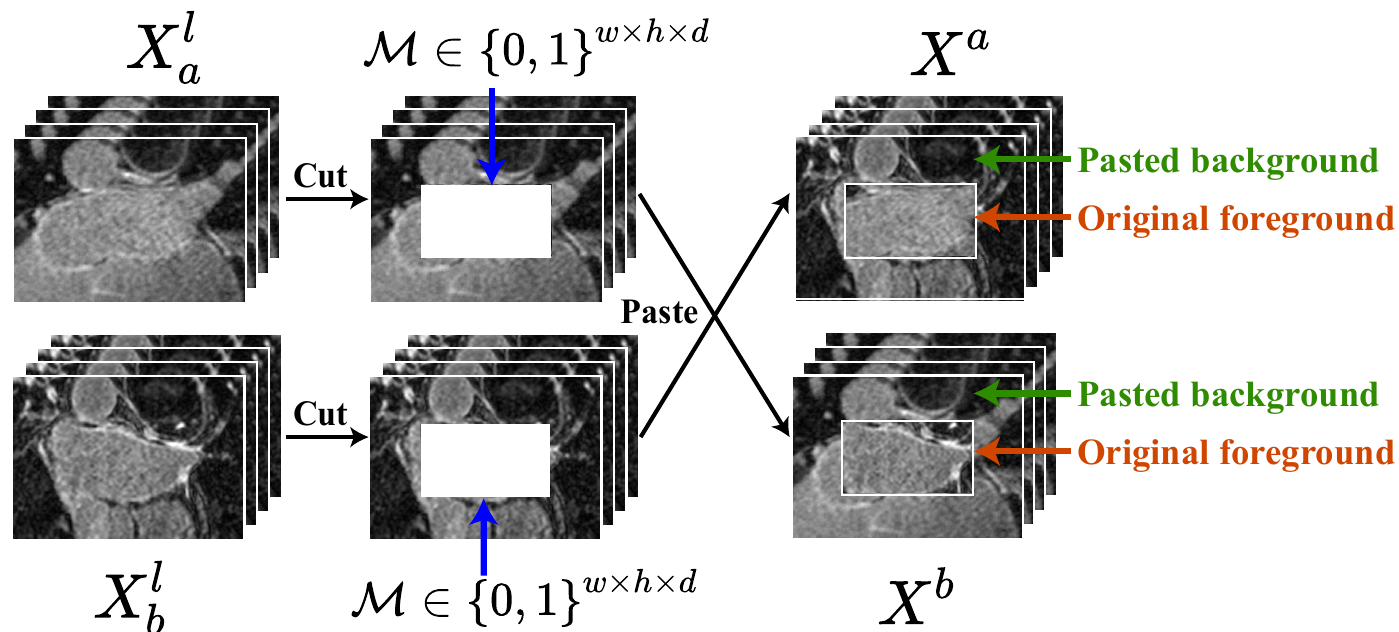}

\caption{{Cut-mix process of labeled data. The enhanced images exchange foreground and background regions. The size of the zero-valued region in $\mathcal{M}$ is $\beta w \times \beta h \times \beta d$.}}

\label{fig:cutmix}
\end{figure}

According to our cross-view bidirectional modeling scheme, {the teacher model generates foreground predictions ${Q}_\text{fg}^{a}$ and ${Q}_\text{fg}^{b}$, background predictions ${Q}_\text{bg}^{a}$ and ${Q}_\text{bg}^{b}$, {and mixed predictions ${Q}_\text{M}^{a}$ and ${Q}_\text{M}^{b}$ via the mixing layer} (detailed in Subsection~\ref{student model}). Bidirectional supervisory optimization is applied utilizing the loss functions $\mathcal{L}_\text{fg}$, $\mathcal{L}_\text{bg}$, and $\mathcal{L}_\text{M}$:}
\begin{equation}
\mathcal{L}_\text{fg} = \mathcal{L}_\text{seg}({Q}_\text{fg}^{a}, {Y}_\text{fg}^{a}) \odot \mathcal{M} + \mathcal{L}_\text{seg}({Q}_\text{fg}^{b}, {Y}_\text{fg}^{b}) \odot (1 - \mathcal{M}),
\label{con:lfg}
\end{equation}
\begin{equation}
\mathcal{L}_\text{bg} = \mathcal{L}_\text{seg}({Q}_\text{bg}^{a}, {Y}_\text{bg}^{a}) \odot \mathcal{M} + \mathcal{L}_\text{seg}({Q}_\text{bg}^{b}, {Y}_\text{bg}^{b}) \odot (1 - \mathcal{M}),
\label{con:lbg}
\end{equation}
\begin{equation}
\mathcal{L}_\text{M} = \mathcal{L}_\text{seg}({Q}_\text{M}^{a}, {Y}_\text{fg}^{a}) \odot \mathcal{M} + \mathcal{L}_\text{seg}({Q}_\text{M}^{b}, {Y}_\text{fg}^{b}) \odot (1 - \mathcal{M}),
\label{con:lS}
\end{equation}
{where $\mathcal{L}_\text{fg}$ and $\mathcal{L}_\text{bg}$ represent the foreground and background loss, respectively. $\mathcal{L}_\text{M}$ denotes the mixed loss.} $\mathcal{L}_\text{seg}$ is a linear combination of Dice loss and Cross-entropy loss. ${Y}^{a}_\text{fg}$ and ${Y}^{b}_\text{fg}$ denote the ground truth labels for foreground modeling, while ${Y}_\text{bg}^{a}$ and ${Y}_\text{bg}^{b}$ are background labels generated by Eq.~(\ref{con:bg1}) or Eq.~(\ref{con:bg2}). 

\begin{figure}[t]
\centering

\includegraphics[width=0.93\columnwidth]{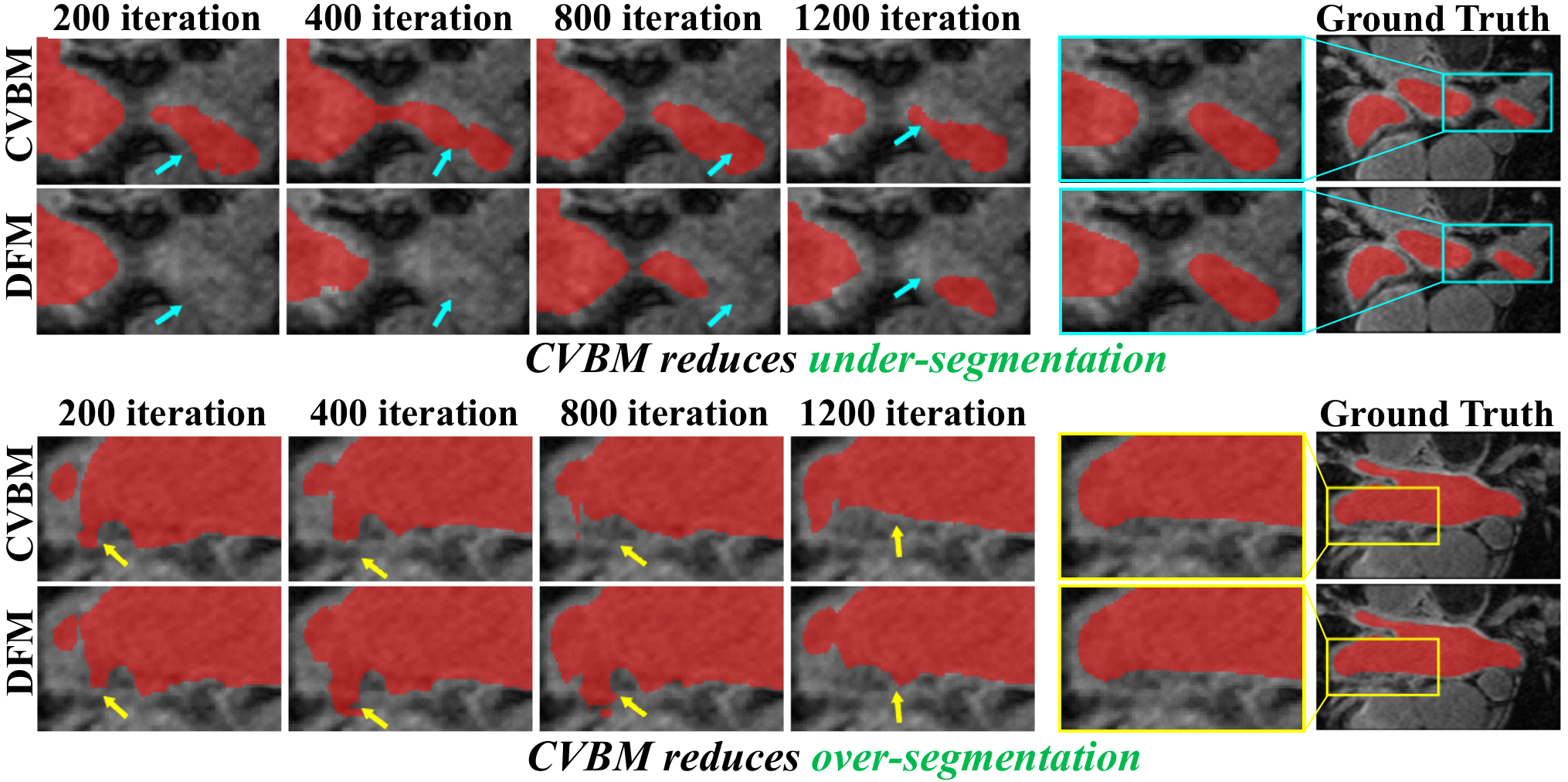}

\caption{Foreground pseudo-labels of the LA dataset during early training stages (200, 400, 800, 1200). \textbf{DFM indicates Dual Foreground Modeling scheme}. Blue rectangles represent under-segmentation, while yellow rectangles indicate over-segmentation. All predictions are derived from the single modeling branch, which employs a conventional VNet architecture.}

\label{fig:epoch}
\end{figure}

\subsubsection{Training Process of Student Model}\label{student model}
The pre-trained teacher model generates foreground and background pseudo-labels (\ie, $P_\text{fg}$ and $P_\text{bg}$) for unlabeled data. The foreground pseudo-labels are illustrated in Fig.~\ref{fig:epoch}. With background modeling, the teacher model achieves accurate target localization within 200 iterations. Additionally, background modeling assists the foreground model in correcting under-segmentation and over-segmentation within 800 iterations. Building on this pre-training, background modeling guides the foreground model to iteratively improve the prediction confidence of uncertain regions, enabling the teacher model to generate more accurate pseudo-labels.
{These pseudo-labels are combined with ground truth labels (${Y}_\text{fg}$ and ${Y}_\text{bg}$) through a cut-mix operation, producing augmented labels ($\hat{Y}_\text{fg}$ and $\hat{Y}_\text{bg}$). This process is formally expressed as:}
\begin{equation}
\hat{Y}_\text{fg} = {Y}_\text{fg} \odot \mathcal{M} + {P}_\text{fg} \odot (1 - \mathcal{M}),
\label{con:bcp1}
\end{equation}
\begin{equation}
\hat{Y}_\text{bg} = {Y}_\text{bg} \odot (1 - \mathcal{M}) + {P}_\text{bg} \odot \mathcal{M},
\label{con:bcp2}
\end{equation}
{where $\hat{Y}_\text{fg}$ and $\hat{Y}_\text{bg}$ provide supervision for training the student model. Each augmented label combines ground truth and pseudo-label.}

\begin{figure*}[t]
\centering

\includegraphics[width=0.95\textwidth]{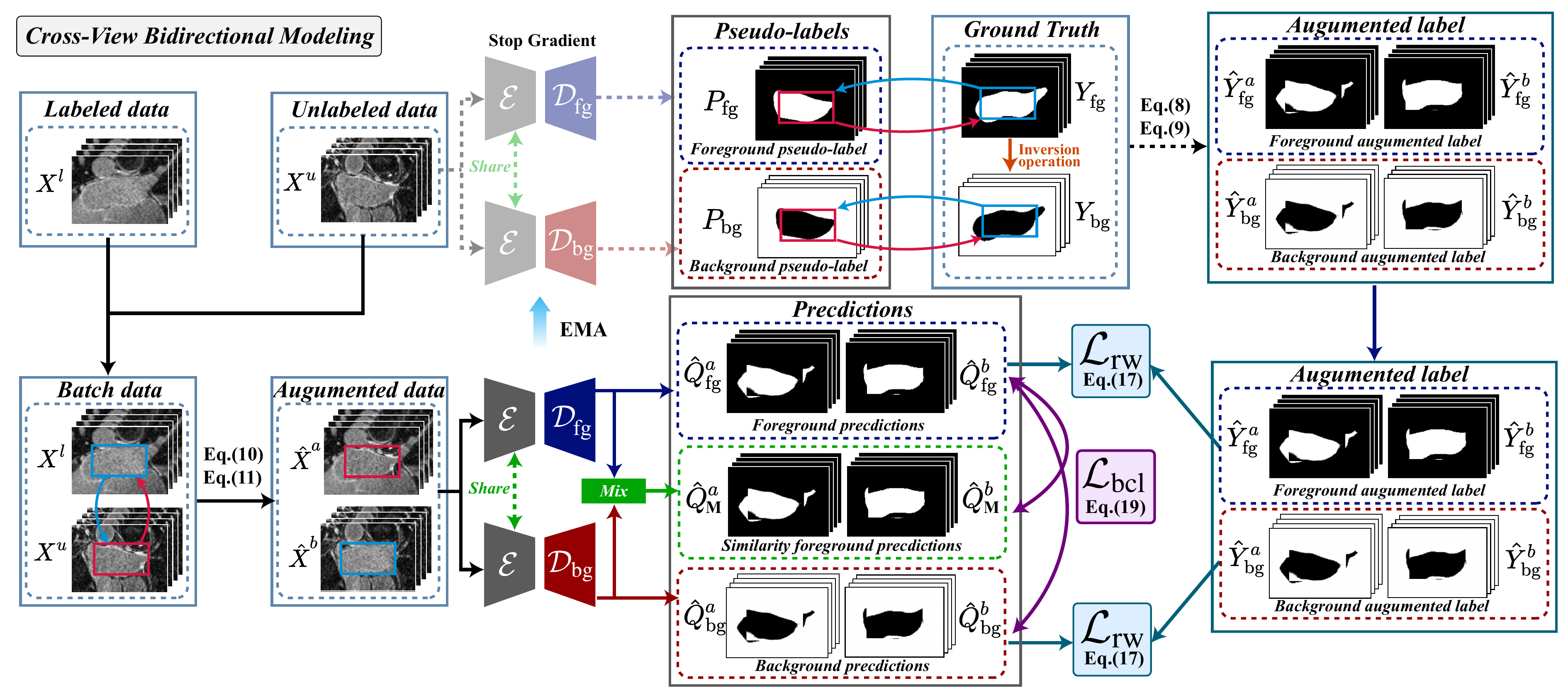}

\caption{Framework of the Cross-view Bidirectional Modeling (CVBM) scheme. The teacher network is pre-trained using labeled data and generates pseudo-labels (${P}_\text{fg}$, ${P}_\text{bg}$) for unlabeled data. During student network training, it processes cut-and-paste inputs (${{{\hat{X}}}^{a}}$, ${{{\hat{X}}}^{b}}$) and performs three core tasks: foreground modeling, background modeling, and mixing, generating the respective predictions $\hat{Q}_\text{fg}^a$ and $\hat{Q}_\text{fg}^b$, $\hat{Q}_\text{bg}^a$ and  $\hat{Q}_\text{bg}^b$, $\hat{Q}_\text{M}^a$ and $\hat{Q}_\text{M}^b$, respectively.. Optimization involves minimizing the Region-wide loss ($\mathcal{L}_\text{rw}$) and the Bidirectional Consistency Loss ($\mathcal{L}_\text{bcl}$).}

\label{fig:banet}
\end{figure*}

{The augmented inputs for the student model are generated by combining labeled data ${X}^{l}$ and unlabeled data ${X}^{u}$:}
\begin{equation}
\hat{X}^{a} = {X}^{l} \odot \mathcal{M} + {X}^{u} \odot (1 - \mathcal{M}),
\label{con:y_fg}
\end{equation}
\begin{equation}
\hat{X}^{b} = {X}^{l} \odot (1 - \mathcal{M}) + {X}^{u} \odot \mathcal{M},
\label{con:y_bg}
\end{equation}
{where $\hat{X}^{a}$ and $\hat{X}^{b}$ represent the augmented data, each containing regions from both labeled and unlabeled data.}

{During modeling, a shared encoder extracts features from $\hat{X}^{a}$ and $\hat{X}^{b}$. For clarity, we utilize $\hat{X}^{a}$ as an example, as the process for $\hat{X}^{b}$ is identical. Two specialized decoders then learn foreground and background information from the shared encoder. The blue decoder, $\mathcal{D}_\text{fg}$, extracts foreground information $\hat{Q}_\text{fg}^{a} \in \mathbb{R}^{w \times h \times d}$, while the yellow decoder, $\mathcal{D}_\text{bg}$, extracts background information $\hat{Q}_\text{bg}^{a} \in \mathbb{R}^{w \times h \times d}$. This cross-view decoder architecture encourages the shared encoder to generate more discriminative features, enhancing the confidence of foreground predictions.}

To fully leverage cross-view modeling, {a mixing layer interconnects the two decoders, generating another foreground prediction $\hat{Q}_\text{M}^{a}$, designed to be similar to $\hat{Q}_\text{fg}^{a}$. This operation is formally expressed as:}
\begin{equation}
\hat{Q}_\text{M}^{a} = \psi(\text{concat}(\hat{Q}_\text{fg}^{a}, \hat{Q}_\text{bg}^{a})),
\label{con:qs}
\end{equation}
{where $\text{concat}(\cdot)$ denotes concatenation along the channel dimension, and $\psi$ represents a $1 \times 1 \times 1$ convolution to ensure dimensional alignment among $\hat{Q}_\text{M}^{a}$, $\hat{Q}_\text{fg}^{a}$, and $\hat{Q}_\text{bg}^{a}$. $\hat{Q}_\text{M}^{a}$ is a foreground prediction influenced by background modeling, playing a pivotal role in bidirectional consistency optimization.}

\subsection{Bidirectional Consistency Optimization}\label{sub:bidirectional consistency}
{The ground truth values of the foreground and background are inherently complementary, as their probabilities sum to 1. This complementary relationship suggests a bidirectional optimization potential between foreground and background modeling. To exploit this, we introduce a bidirectional consistency optimization scheme to supervise cross-view feature learning. This scheme imposes constraints on both foreground and background segmentation outputs through two key loss functions: {the Region-wide Loss ($\mathcal{L}_\text{rw}$)} and the Bidirectional Consistency Loss ($\mathcal{L}_\text{bcl}$). The Region-wide Loss refines segmentation results for both foreground and background branches, while the Bidirectional Consistency Loss enforces consistency between foreground predictions and background-guided predictions. The following sections analyze each loss function and their contributions to the optimization process.}

\subsubsection{Region-wide Loss}
{Different from conventional foreground optimization functions, $\mathcal{L}_\text{rw}$ extends the optimization scope to include the background region, enabling pixel-wise supervision across the entire image. Specifically, $\mathcal{L}_\text{rw}$ consists of two components: a labeled part and an unlabeled part. Each component supervises predictions from both foreground modeling (\ie, $\hat{Q}_\text{fg}^{a}$ and $\hat{Q}_\text{fg}^{b}$) and background modeling (\ie, $\hat{Q}_\text{bg}^{a}$ and $\hat{Q}_\text{bg}^{b}$).} {The labeled component is defined as:}
\begin{equation}
\mathcal{L}_\text{fg}^{l} = \mathcal{L}_\text{seg}(\hat{Q}_\text{fg}^{a}, \hat{Y}_\text{fg}^{a}) \odot \mathcal{M} + \mathcal{L}_\text{seg}(\hat{Q}_\text{fg}^{b}, \hat{Y}_\text{fg}^{b}) \odot (1 - \mathcal{M}),
\label{con:lfg}
\end{equation}
\begin{equation}
\mathcal{L}_\text{bg}^{l} = \mathcal{L}_\text{seg}(\hat{Q}_\text{bg}^{a}, \hat{Y}_\text{bg}^{a}) \odot \mathcal{M} + \mathcal{L}_\text{seg}(\hat{Q}_\text{bg}^{b}, \hat{Y}_\text{bg}^{b}) \odot (1 - \mathcal{M}).
\label{con:lbg}
\end{equation}
{{where $\mathcal{L}_\text{fg}^{l}$ and $\mathcal{L}_\text{bg}^{l}$ represent the foreground and background loss functions for the labeled data, respectively.} Similarly, the unlabeled component is formulated as:}
\begin{equation}
\mathcal{L}_\text{fg}^{u} = \mathcal{L}_\text{seg}(\hat{Q}_\text{fg}^{a}, \hat{Y}_\text{fg}^{a}) \odot (1 - \mathcal{M}) + \mathcal{L}_\text{seg}(\hat{Q}_\text{fg}^{b}, \hat{Y}_\text{fg}^{b}) \odot \mathcal{M},
\label{con:ufg}
\end{equation}
\begin{equation}
\mathcal{L}_\text{bg}^{u} = \mathcal{L}_\text{seg}(\hat{Q}_\text{bg}^{a}, \hat{Y}_\text{bg}^{a}) \odot (1 - \mathcal{M}) + \mathcal{L}_\text{seg}(\hat{Q}_\text{bg}^{b}, \hat{Y}_\text{bg}^{b}) \odot \mathcal{M}.
\label{con:ubg}
\end{equation}
{{where $\mathcal{L}_\text{fg}^{u}$ and $\mathcal{L}_\text{bg}^{u}$ represent the foreground and background loss functions for the unlabeled data, respectively.} The Region-wide Loss is then defined as:}
\begin{equation}
\mathcal{L}_\text{rw} = \mathcal{L}_\text{fg}^{l} + \mathcal{L}_\text{bg}^{l} + \alpha (\mathcal{L}_\text{fg}^{u} + \mathcal{L}_\text{bg}^{u}),
\label{con:rw}
\end{equation}
{where $\alpha$ balances the labeled and unlabeled losses.}

\subsubsection{Bidirectional Consistency Loss}
{To enhance feature similarity within foreground regions and improve discriminability between foreground and background features, we introduce the Bidirectional Consistency Loss ($\mathcal{L}_\text{bcl}$). This loss optimizes foreground modeling, background modeling, and mixing layer predictions within the student network. It consists of two consistency terms:}
{1) \textit{Direct consistency}: For foreground predictions ($\hat{Q}_\text{fg}^{a}$ and $\hat{Q}_\text{M}^{a}$), $\mathcal{L}_\text{bcl}$ establishes direct consistency to reduce intra-class spacing.}
{2) \textit{Inverse consistency}: For background predictions ($\hat{Q}_\text{fg}^{a}$ and $\hat{Q}_\text{bg}^{a}$), $\mathcal{L}_\text{bcl}$ implements inverse consistency to increase inter-class spacing:}
\begin{equation}
\mathcal{L}_\text{bcl}^{a} = \underbrace{\mathcal{L}_\text{mse}(\hat{Q}_\text{M}^{a}, \hat{Q}_\text{fg}^{a})}_{\text{Direct consistency}} + \underbrace{\mathcal{L}_\text{mse}((1 - \hat{Q}_\text{bg}^{a}), \hat{Q}_\text{fg}^{a})}_{\text{Inverse consistency}},
\label{con:bcla}
\end{equation}
{where $\mathcal{L}_\text{mse}$ denotes the Mean Squared Error loss. Similarly, $\mathcal{L}_\text{bcl}^{b}$ is derived, and the overall $\mathcal{L}_\text{bcl}$ is defined as:}
\begin{equation}
\mathcal{L}_\text{bcl} = \mathcal{L}_\text{bcl}^{a} + \mathcal{L}_\text{bcl}^{b}.
\label{con:bcl}
\end{equation}

\begin{figure}[t]
\centering
\includegraphics[width=0.95\columnwidth]{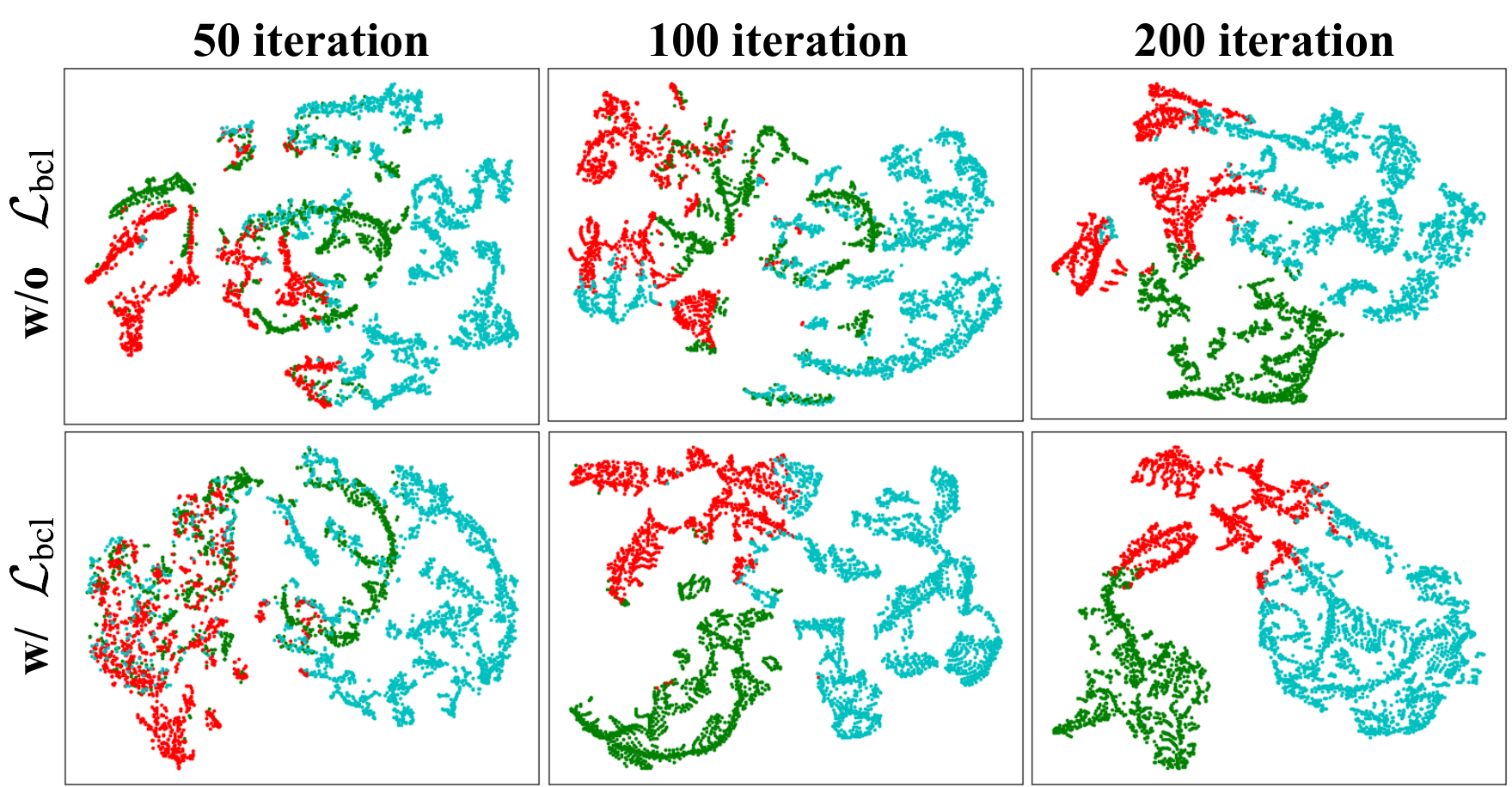}

\caption{{t-SNE visualization of the ACDC dataset with 10\% labeled data at different training stages (50, 100, and 200 iterations). w/o denotes features extracted before the classification layer. Different colors represent distinct categories in the ACDC dataset.}}
\label{fig:TSNE}

\end{figure}

{$\mathcal{L}_\text{bcl}$ leverages $\hat{Q}_\text{M}^{a}$ and $\hat{Q}_\text{bg}^{a}$ to jointly constrain foreground feature learning. As shown in Fig.~\ref{fig:TSNE}, bidirectional consistency improves intra-cluster cohesion and inter-cluster separation in multi-target segmentation, enabling the network to generate discriminative features and enhance segmentation accuracy.}

{In summary, the bidirectional consistency optimization scheme is formulated as:}
\begin{equation}
\mathcal{L}_\text{total} = \mathcal{L}_\text{rw} + \lambda \mathcal{L}_\text{bcl},
\label{con:total}
\end{equation}
{where $\lambda$ balances the contributions of $\mathcal{L}_\text{rw}$ and $\mathcal{L}_\text{bcl}$.}

\subsubsection{Inference}
{In our framework, background modeling serves as an auxiliary mechanism to enhance foreground modeling performance. During inference, only the foreground branch of the student model is utilized, \textbf{ensuring no additional computational overhead.}}
\subsubsection{Theoretical Analysis}
{We provide theoretical insights into background-assisted modeling in Appendix B. Theorem 1 demonstrates that cross-view modeling exhibits higher prediction confidence compared to traditional foreground-oriented training methods. Furthermore, in Theorem 2, we further establish that foreground and background modeling achieve dynamic bidirectional optimization within the cross-view framework. Additional proof details could be found in our Appendix B.}}

\begin{algorithm}
\caption{Training Pipeline of CVBM}
\label{alg:CVBM}
\KwIn{Labeled samples $D^{l} = \{(X_{i}^{l}, Y_{i}^{l})\}_{i=1}^{N_{L}}$, unlabeled samples $D^{u} = \{X_{i}^{u}\}_{i=1}^{N_{U}}$}
\KwOut{Shared encoder $\mathcal{E}$, foreground decoder $\mathcal{D}_\text{fg}$, and background decoder $\mathcal{D}_\text{bg}$. Only $\mathcal{E}$ and $\mathcal{D}_\text{fg}$ are utilized during inference.}
{
\For{batched data $\{(X^{l}, Y^{l})\}$, $X^{u}$} {
    Generate complementary background labels $Y_\text{bg}^{l}$ utilizing Eq.~(\ref{con:bg1}) or Eq.~(\ref{con:bg2});\\
    Generate augmented data $\hat{X}^{a}, \hat{X}^{b}$ utilizing Eq.~(\ref{con:y_fg}) and Eq.~(\ref{con:y_bg});
    
    {\tcp{Teacher model}}
    Generate pseudo-labels $P_\text{fg}$ and $P_\text{bg}$:
    $P_\text{fg} \leftarrow \mathcal{D}_\text{fg}(\mathcal{E}(X^{u}))$;
    $P_\text{bg} \leftarrow \mathcal{D}_\text{bg}(\mathcal{E}(X^{u}))$;
    
    Generate augmented foreground labels $\hat{Y}_\text{fg}^{a}, \hat{Y}_\text{fg}^{b}$ utilizing $(Y^{l}, P_\text{fg})$;\\
    Generate augmented background labels $\hat{Y}_\text{bg}^{a}, \hat{Y}_\text{bg}^{b}$ utilizing $(Y_\text{bg}^{l}, P_\text{bg})$;
    
    {\tcp{Student model}}
    Extract foreground predictions $\hat{Q}_\text{fg}^{a} \leftarrow \mathcal{D}_\text{fg}(\mathcal{E}(\hat{X}^{a}))$;\\
    Extract background predictions $\hat{Q}_\text{bg}^{a} \leftarrow \mathcal{D}_\text{bg}(\mathcal{E}(\hat{X}^{a}))$;\\
    Generate mixed predictions $\hat{Q}_\text{M}^{a} \leftarrow \psi(\text{concat}(\hat{Q}_\text{fg}^{a}, \hat{Q}_\text{bg}^{a}))$;
    
    Generate $\hat{Q}_\text{fg}^{b}, \hat{Q}_\text{bg}^{b}, \hat{Q}_\text{M}^{b}$ for $\hat{X}^{b}$ following the same procedure as for $\hat{X}^{a}$;
    
    Calculate Region-wide Loss $\mathcal{L}_\text{rw}$ utilizing Eq.~(\ref{con:rw});\\
    Calculate Bidirectional Consistency Loss $\mathcal{L}_\text{bcl}$ utilizing Eq.~(\ref{con:bcl});
    
    Optimize student model parameters ($\mathcal{E}$, $\mathcal{D}_\text{fg}$, $\mathcal{D}_\text{bg}$) utilizing SGD;\\
    Update teacher model parameters ($\mathcal{E}$, $\mathcal{D}_\text{fg}$, $\mathcal{D}_\text{bg}$) utilizing EMA.
}}
\end{algorithm}

In summary, our learning scheme for CVBM is {articulated in} Algorithm \ref{alg:CVBM}. CVBM introduces a novel modeling perspective by {establishing} a cross-view bidirectional modeling approach. The shared encoder $\mathcal{E}$ {captures} global information from both foreground and background modeling ($\mathcal{D}_\text{fg}$ and $\mathcal{D}_\text{bg}$, respectively). As a result, for a single instance, our modeling scheme covers every voxel of the medical image, {enabling comprehensive learning of the input data semantics}.
Additionally, our proposed bidirectional consistency optimization scheme, {comprising} $\mathcal{L}_\text{rw}$ and $\mathcal{L}_\text{bcl}$, establishes bidirectional consistency between $\mathcal{D}_\text{fg}$ and $\mathcal{D}_\text{bg}$. This provides global supervision for CVBM's prediction results. By incorporating background modeling, the foreground modeling reduces uncertain predictions, thereby enhancing the predictive confidence of the foreground model.

\section{Experiments}
\subsection{Datasets}
\subsubsection{LA Dataset} It serves as benchmark dataset for the 2018 Atrial Segmentation Challenge~\cite{xiongGlobalBenchmarkAlgorithms2021}, comprising 100 3D gadolinium-enhanced MR imaging scans with expert annotations. 
The dataset has an isotropic resolution of 0.625 × 0.625 × 0.625 mm. We adopted a standardized setup~\cite{luoEfficientSemisupervisedGross2021, wangMCFMutualCorrection2023}, utilizing 80 samples for training and 20 samples for testing. 
\subsubsection{NIH-Pancreas Dataset} It consists of 82 3D abdominal contrast-enhanced CT scans, was publicly released by National Institutes of Health Clinical Center~\cite{clarkCancerImagingArchive2013}. The acquisition of these data is conducted on Philips and Siemens MDCT scanners, with a fixed resolution of 512 × 512 and varying thicknesses ranging from 1.5 to 2.5 mm. We utilized a uniform data splitting approach~\cite{wuMutualConsistencyLearning2022, wangMCFMutualCorrection2023}, allocating 62 samples for training and remaining 20 samples for testing. 
\subsubsection{ACDC Dataset} It was collected from clinical examinations acquired at University Hospital of Dijon~\cite{bernardDeepLearningTechniques2018}. It consists of cardiac MR imaging samples collected from 100 patients. For data management, we employed a consistent data splitting~\cite{wuExploringSmoothnessClassSeparation2022, wuMutualConsistencyLearning2022}, allocating 70, 10, and 20 patient scans for training, validation, and test sets, respectively. 
{\subsubsection{HRF Dataset}  It is a dataset from Tomas Kubena's Ophthalmology Clinic in Czech Republic~\cite{odstrcilikRetinalVesselSegmentation2013a}, containing 45 color fundus images (15 healthy, 15 diabetic retinopathy, 15 glaucoma) with expert-annotated vessel segmentation labels. About data splitting, we randomly selected 27 images for training and 18 images for testing.} 

\begin{table*}[htbp]
 \caption{Comparisons with SOTA semi-supervised segmentation methods on LA dataset. ↑ indicates the higher the better, ↓ indicates the lower the better. The highest result is \textbf{bolded}, while the second highest result is \underline{underlined}.}
 
 \centering
 \setlength{\tabcolsep}{1.5 mm}{
 \renewcommand{\arraystretch}{0.94}
 \begin{tabular}{c|c|c|cccc|cc|c}
    \toprule[1pt]
\multirow{2}[4]{*}{Type} & \multirow{2}[4]{*}{Methods} & \multicolumn{1}{c|}{Scans Used} & \multicolumn{4}{c|}{Metrics} & \multicolumn{2}{c|}{{Inference Cost}} & \multirow{2}[4]{*}{\makecell[c]{{p-value↓}\\{$(<0.05)$}}} \\
\cmidrule{3-9}          &       & \multicolumn{1}{c|}{Label/Unlabel} & \multicolumn{1}{c}{DSC↑(\%)} & \multicolumn{1}{c}{Jaccard↑(\%)} & \multicolumn{1}{c}{95HD↓(voxel)} & \multicolumn{1}{c}{ASD↓(voxel)} & {Parameters} & {FLOPs} & \\
    \midrule
    Fully-supervised & VNet (Lower Bound) & 4/0   & 52.55 & 39.60 & 47.05 & 9.87 & {9.45M} & {47.17G} & {$\surd$} \\
    \midrule
    \multirow{8}[1]{*}{Semi-supervised} & UA-MT~\cite{yuUncertaintyAwareSelfensemblingModel2019} & 4/76  & 82.26  & 70.98  & 13.71  & 3.82 & {9.45M} & {47.17G} &  {$\surd$} \\
          & SASSNet~\cite{liShapeAwareSemisupervised3D2020} & 4/76  & 81.60  & 69.63 & 16.16 & 3.58 & {9.45M} & {47.17G} & {$\surd$} \\
          & DTC~\cite{luoSemisupervisedMedicalImage2021}   & 4/76  & 81.25 & 69.33 & 14.90 & 3.99 & {9.45M} & {47.17G} & {$\surd$} \\
          & URPC~\cite{luoEfficientSemisupervisedGross2021}  & 4/76  & 82.48 & 71.35 & 14.65 & 3.65 & {9.45M} & {47.17G} & {$\surd$} \\
          & MC-Net~\cite{wuSemisupervisedLeftAtrium2021} & 4/76  & 83.59 & 72.36 & 14.07 & 2.70 & {9.45M} & {47.17G} & {$\surd$} \\
          & SS-Net~\cite{wuExploringSmoothnessClassSeparation2022} & 4/76  & 86.33 & 76.15 & 9.97  & 2.31 & {9.45M} & {47.17G} & {$\surd$} \\
          & BCP~\cite{baiBidirectionalCopyPasteSemiSupervised2023}   & 4/76  & \underline{88.02} & \underline{78.72} & \underline{7.90}  & \underline{2.15} & {9.45M} & {47.17G} & {$\surd$} \\
          & \textbf{CVBM (Ours)} & \textbf{4/76} & \textbf{89.50}  & \textbf{81.07} & \textbf{5.78}  & \textbf{2.10} & {9.45M} & {47.17G} & {-} \\
    \midrule
    \midrule
    Fully-supervised & VNet (Lower Bound) & 8/0   & 82.74 & 71.72 & 13.35 & 3.26 & {9.45M} & {47.17G} & {$\surd$} \\
    \midrule
    \multirow{8}[1]{*}{Semi-supervised} 
          & UA-MT~\cite{yuUncertaintyAwareSelfensemblingModel2019} & 8/72  & 87.79 & 78.39 & 8.68  & 2.12 & {9.45M} & {47.17G} & {$\surd$} \\
          & SASSNet~\cite{liShapeAwareSemisupervised3D2020} & 8/72  & 87.54 & 78.05 & 9.84  & 2.59 & {9.45M} & {47.17G} & {$\surd$} \\
          & DTC~\cite{luoSemisupervisedMedicalImage2021}   & 8/72  & 87.51 & 78.17 & 8.23  & 2.36 & {9.45M} & {47.17G} & {$\surd$} \\
          & URPC~\cite{luoEfficientSemisupervisedGross2021}  & 8/72  & 86.92 & 77.03 & 11.13 & 2.28 & {9.45M} & {47.17G} & {$\surd$} \\
          & MC-Net~\cite{wuSemisupervisedLeftAtrium2021} & 8/72  & 87.62 & 78.25 & 10.03 & 1.82 & {9.45M} & {47.17G} & {$\surd$} \\
          & SS-Net~\cite{wuExploringSmoothnessClassSeparation2022} & 8/72  & 88.55 & 79.62 & 7.49  & 1.90 & {9.45M} & {47.17G} & {$\surd$} \\
          & BCP~\cite{baiBidirectionalCopyPasteSemiSupervised2023}   & 8/72  & \underline{89.62} & \underline{81.31} & \underline{6.81}  & \underline{1.76} & {9.45M} & {47.17G} & {$\surd$} \\
          & \textbf{CVBM (Ours)} & \textbf{8/72}  & \textbf{91.19} & \textbf{83.87} & \textbf{5.45}  & \textbf{1.61} & {9.45M} & {47.17G} & {-} \\
    \midrule
    \midrule
      \multirow{4}[2]{*}{Fully-supervised} &  SAM~\cite{Kirillov_2023_ICCV} & -  & 72.52 & 59.60 & 13.76  & 4.28 & {93.74M} & {370.63G} & {$\surd$} \\
    & SAM\_Med3D~\cite{wangSAMMed3D2023} & -  & 75.19 & 62.10 & 11.25  & 3.18 & {100.51M} & {89.85G} & {$\surd$} \\
    & VNet (Upper Bound) & 80/0  & 91.47 & 84.36 & 5.48  & 1.50 & {9.45M} & {47.17G} & {$\surd$} \\
     & {CVBM (Ours)} & {80/0}  & {92.02} & {85.28} & {5.07}  & {1.46} & {9.45M} & {47.17G} & {-} \\
    \bottomrule[1pt]
    \end{tabular}%
    }
  \label{tab:LA}%

\end{table*}

\begin{figure*}[t]
\centering
{\includegraphics[width=0.95\textwidth]{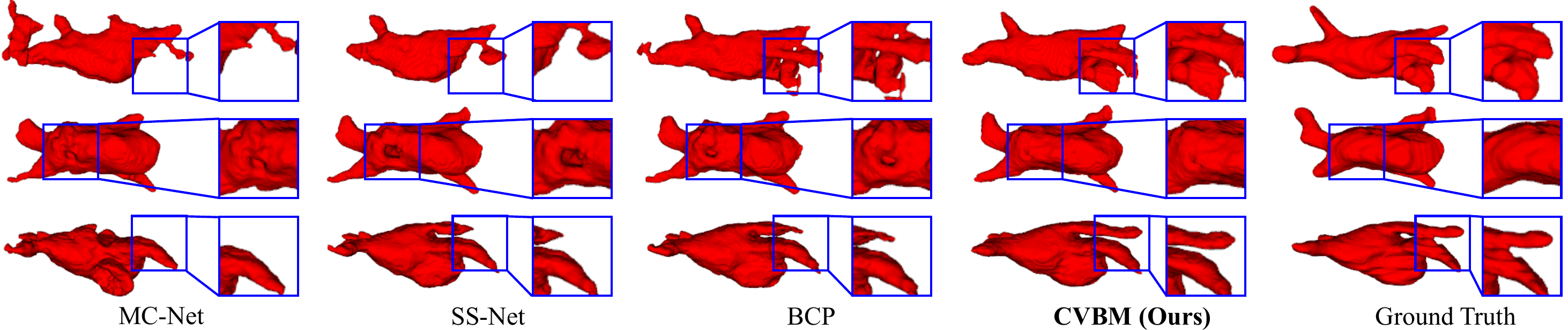}}% 蓝色框

\caption{{3D visualization results of LA dataset with 8/72 labeled data.  CVBM reduces the occurrence of discretization errors, producing smoother and more accurate segmentation surface. Best viewed by zoom-in on screen. }}

\label{fig:3D}
\end{figure*}
\subsection{Implementation Details}
Our scheme is implemented in PyTorch 1.12.1 and trained iteratively on one NVIDIA 3090 GPU. The total batch size is set to 8, including 4 labeled and 4 unlabeled images. Segmentation tasks are optimized {utilizing} SGD with an initial learning rate of 0.01. Hyper-parameters are set as follows: {Following~\cite{yuUncertaintyAwareSelfensemblingModel2019,luoSemisupervisedMedicalImage2021,wuSemisupervisedLeftAtrium2021,baiBidirectionalCopyPasteSemiSupervised2023},} $\lambda$ is determined {utilizing}  a time-dependent Gaussian preheating function $\lambda(t) = 0.1 \times e^{-5(1-t/t_{\max})^2}$~\cite{laineTemporalEnsemblingSemiSupervised2017}, where $t$ and $t_{\max}$ are current and total training steps. $\beta$ is set to 2/3~\cite{baiBidirectionalCopyPasteSemiSupervised2023}, and $\alpha$ is empirically set to 0.5. Following~\cite{wuExploringSmoothnessClassSeparation2022}, all 3D volumes were normalized with zero mean and unit variance. Sub-volumes of size 112$\times$112$\times$80 (LA) and 96$\times$96$\times$96 (Pancreas-CT) were randomly cropped as input. Pre-training and self-training iterations were 2k, 15k (LA) and 3k, 15k (Pancreas-CT) respectively. For the ACDC dataset, we followed~\cite{baiBidirectionalCopyPasteSemiSupervised2023, wuMutualConsistencyLearning2022}, extracting 2D patches of size 256$\times$256 as input. Pre-training and self-training iterations were 10k and 30k respectively. During training, data augmentation including rotations and flips was applied. During testing, following~\cite{luoEfficientSemisupervisedGross2021}, we used a sliding window strategy with an 18$\times$18$\times$4 step size for LA dataset and 16$\times$16$\times$16 step size for pancreas dataset. Importantly, for a fair comparison during testing, we utilized only the foreground model, which is a traditional VNet.
\subsection{Comparison with Sate-of-the-Art Methods}
{To comprehensively evaluate the proposed CVBM framework, we conduct extensive comparisons with SOTA methods across four benchmark datasets. Specifically, we employ 3D methods on the LA and Pancreas-CT datasets to validate the framework's effectiveness in 3D medical image segmentation. For 2D scenarios, we utilize the ACDC dataset to assess multi-class segmentation performance and the HRF dataset to demonstrate its applicability in 2D color image analysis.} For quantitative evaluation, we adopted four metrics: Dice similarity coefficient (DSC), Jaccard index (Jaccard), average surface distance (ASD), and the 95th percentile Hausdorff distance (95HD). 
\subsubsection{LA Dataset}
Table \ref{tab:LA} presents the quantitative results for the LA dataset under the 4/76 and 8/72 semi-supervised settings. Additionally, it illustrates the performance of the fully supervised V-Net model trained with 4, 8, and 80 labeled data as reference benchmarks. {It is worth noting that, under the experimental setting with 4 labeled data, UA-MT successfully leveraged the unlabeled data (76 volumes) to improve the fully supervised performance from 52.55\% to 82.26\% {($p < 0.01$)}. While MC-Net further enhanced the performance to 83.59\% {($p < 0.01$)} by constructing mutual consistency among the foreground predictions.} Nevertheless, in the experimental setting with 8 labeled data, the performance advantage of MC-Net diminished. This observation further demonstrates that relying solely on foreground modeling does not guarantee the generation of more discriminative features. In comparison with other competitive methods, our CVBM successfully enhanced semi-supervised segmentation performance to surpass 90\% utilizing only 8 labeled data. This result further demonstrates the advantages of cross-view modeling in SSMIS. Moreover, CVBM exhibits improved performance when compared to models with larger parameters and those trained on extensive datasets, \ie, SAM~\cite{Kirillov_2023_ICCV} and SAM\_Med3D~\cite{wangSAMMed3D2023}. Notably, utilizing 8 labeled data, CVBM surpasses the segmentation upper bound (with 80 labeled data) on 95HD. {Furthermore, CVBM outperforms SOTA algorithms across all semi-supervised settings without incurring additional inference costs.}

Figure \ref{fig:3D} illustrates the 3D segmentation results of CVBM in comparison with SOTA methods. {Within the blue-circled regions, MC-Net, SS-Net and BCP produce discrete mispredictions that manifest as surface protrusions or depressions. This indicates that in challenging areas, the foreground model tend to produce uncertain predictions. During the final voxel-classification stage, these low-confidence regions are removed, leading to inaccurate predictions. In contrast, our CVBM approach generates smoother and more precise boundaries that closely approximate the ground truth. This demonstrates that with the assistance of background modeling, the foreground model mitigates uncertain predictions, producing the predictions that closely match the ground truth.}

\begin{table*}[htbp]
 \caption{{Comparisons with SOTA semi-supervised segmentation methods on Pancreas dataset.}}

 \centering
 \setlength{\tabcolsep}{1.5mm}{
 \renewcommand{\arraystretch}{0.94}
 \begin{tabular}{c|c|c|cccc|cc|c}
    \toprule[1pt]
\multirow{2}[4]{*}{Type} & \multirow{2}[4]{*}{Methods} & \multicolumn{1}{c|}{Scans Used} & \multicolumn{4}{c|}{Metrics} & \multicolumn{2}{c|}{{Inference Cost}} & \multirow{2}[4]{*}{\makecell[c]{{p-value↓}}} \\
\cmidrule{3-9}          &       & \multicolumn{1}{c|}{Label/Unlabel} & \multicolumn{1}{c}{DSC↑(\%)} & \multicolumn{1}{c}{Jaccard↑(\%)} & \multicolumn{1}{c}{95HD↓(voxel)} & \multicolumn{1}{c}{ASD↓(voxel)} & {Parameters} & {FLOPs} & \\
    \midrule
    Fully-supervised & VNet (Lower Bound) & 6/0   & 55.20 & 41.23 & 30.62 & 10.54 & {9.45M} & {41.58G} & {$\surd$} \\
    \midrule
    \multirow{8}[2]{*}{Semi-supervised} & UA-MT~\cite{yuUncertaintyAwareSelfensemblingModel2019} & 6/56  & 66.44 & 52.02 & 17.04 & 3.03 & {9.45M} & {41.58G} & {$\surd$} \\
          & SASSNet~\cite{liShapeAwareSemisupervised3D2020} & 6/56  & 68.97 & 54.29 & 18.83 & 1.96 & {9.45M} & {41.58G} & {$\surd$} \\
          & DTC~\cite{luoSemisupervisedMedicalImage2021}   & 6/56  & 66.58 & 51.79 & 15.46 & 4.16 & {9.45M} & {41.58G} &  {$\surd$} \\
          & MC-Net~\cite{wuSemisupervisedLeftAtrium2021} & 6/56  & 69.07 & 54.36 & 14.53 & 2.28 & {9.45M} & {41.58G} & {$\surd$} \\
          & URPC~\cite{luoEfficientSemisupervisedGross2021}  & 6/56  & 73.53 & 59.44 & 22.57 & 7.85 & {9.45M} & {41.58G} & {$\surd$} \\
          & CauSSL~\cite{miaoCauSSLCausalityinspiredSemisupervised2023} & 6/56  & 72.89 & 58.06 & 14.19 & 4.37 & {9.45M} & {41.58G} & {$\surd$} \\
          & BCP~\cite{baiBidirectionalCopyPasteSemiSupervised2023}   & 6/56  & \underline{82.03} & \underline{69.80} & \underline{5.89}  & \underline{1.96} & {9.45M} & {41.58G} & {$\surd$} \\
          & {\textbf{CVBM (Ours)}} & {\textbf{6/56}}  & {\textbf{83.65}} & {\textbf{72.16}} & {\textbf{4.48}}  & {\textbf{1.30}} & {9.45M} & {41.58G} & {-} \\
    \midrule
    \midrule
    Fully-supervised & VNet (Lower Bound) & 12/0   & 72.38 & 56.78 & 18.12 & 5.41 & {9.45M} & {41.58G} & {$\surd$} \\
    \midrule
    \multirow{8}[2]{*}{Semi-supervised} & UA-MT~\cite{yuUncertaintyAwareSelfensemblingModel2019} & 12/50  & 77.26 & 63.82 & 11.90 & 3.06 & {9.45M} & {41.58G} & {$\surd$} \\
          & SASSNet~\cite{liShapeAwareSemisupervised3D2020} & 12/50  & 77.66 & 64.08 & 10.93 & 3.05 & {9.45M} & {41.58G} & {$\surd$} \\
          & DTC~\cite{luoSemisupervisedMedicalImage2021}   & 12/50  & 78.27 & 64.75 & 8.36  & 2.25 & {9.45M} & {41.58G} & {$\surd$} \\
          & MC-Net~\cite{wuSemisupervisedLeftAtrium2021} & 12/50  & 78.17 & 65.22 & 6.90  & 1.55 & {9.45M} & {41.58G} & {$\surd$} \\
          & URPC~\cite{luoEfficientSemisupervisedGross2021}  & 12/50  & 80.02 & 67.30 & 8.51  & 1.98 & {9.45M} & {41.58G} & {$\surd$} \\
          & CauSSL~\cite{miaoCauSSLCausalityinspiredSemisupervised2023} & 12/50  & 80.92 & 68.26 & 8.11  & 1.53 & {9.45M} & {41.58G} & {$\surd$} \\
          & BCP~\cite{baiBidirectionalCopyPasteSemiSupervised2023}   & 12/50  & \underline{83.03} & \underline{71.25} & \underline{5.22}  & \underline{1.39} & {9.45M} & {41.58G} & {$\surd$} \\
          & {\textbf{CVBM (Ours)}} & {\textbf{12/50}} & {\textbf{84.57}} & {\textbf{73.56}} & {\textbf{3.90}}  & {\textbf{1.21}} & {9.45M} & {41.58G} & {-} \\
    \midrule
    \midrule
     \multirow{4}[2]{*}{Fully-supervised} &  SAM~\cite{Kirillov_2023_ICCV} & -  &61.51 & 46.68 & 14.17  & 9.95 & {93.74M} & {370.63G} & {$\surd$} \\
    &SAM\_Med3D~\cite{wangSAMMed3D2023} & -  & 70.40 & 57.54 & 12.68  & 7.51 & {100.51M} & {89.85G} & {$\surd$} \\
    & VNet (Upper Bound) & 62/0  & 83.89 & 71.91 & 5.08  & 2.00 & {9.45M} & {41.58G} & {$\surd$} \\
     & {CVBM (Ours)} & {62/0}  & {85.52} & {73.84} & {3.61}  & {1.12} & {9.45M} & {41.58G} & {-} \\
    \bottomrule[1pt]
    \end{tabular}%
    }%
   
 \label{tab:Pancreas}%
\end{table*}

\begin{figure*}[t]
\centering
{\includegraphics[width=0.95\textwidth]{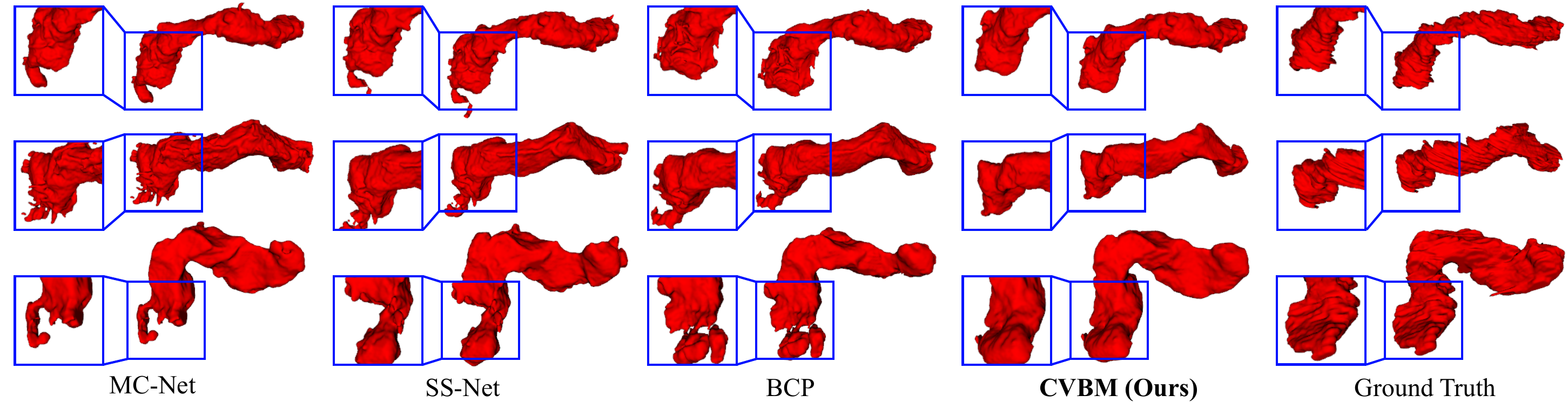}}

\caption{{3D visualization results of Pancreas dataset with 12/50 labeled data. CVBM mitigates both over-segmentation and under-segmentation, and the 3D surface is more closely approximates the ground truth. Best viewed by zoom-in on screen.} }
\label{fig:3Dpancres}

\end{figure*}

\subsubsection{Pancreas-CT Dataset}
The pancreas presents {anatomical variability} with complex surrounding structures occupying substantial image portions, making its segmentation more challenging than organs with simpler boundaries (\eg, liver and spleen). As Table \ref{tab:Pancreas} demonstrates, SASSNet {enhanced pancreatic shape awareness through global constraints}, improving DSC from 55.20\% to 68.79\% with 6 labeled samples. URPC further increased performance to 72.89\% via {multi-level feature consistency}. Despite these advances, previous SOTA methods failed to exceed fully supervised performance. Notably, our CVBM maintains robust performance in both 6/56 and 12/50 labeled data settings, surpassing the upper bound (62 labeled data) across all evaluation metrics when using just 12/50 labeled samples. This demonstrates that background modeling assistance enables effective prediction even with complex pancreatic backgrounds. {Compared to SAM and SAM\_Med3D, our approach achieves 23.06\% and 14.7\% performance improvements respectively {($p< 0.05$)}, while utilizing {substantially fewer parameters}.}

Fig. \ref{fig:3Dpancres} presents 3D visualization results from the pancreas dataset. Pure foreground modeling networks (MC-Net, SS-Net, and BCP) exhibit under-segmentation errors in specific regions. In contrast, CVBM's background modeling effectively refines these mispredicted areas, yielding reconstructed morphology closely resembling ground truth. This underscores the efficacy of background modeling in enhancing segmentation accuracy, particularly for challenging datasets with complex background structures. {By leveraging cross-view modeling, CVBM produces  precise and reliable organ segmentation.}

\begin{table*}[htbp]
 \caption{{Comparisons with SOTA semi-supervised segmentation methods on ACDC dataset.}}

 \centering
 \setlength{\tabcolsep}{1.5mm}{
 \renewcommand{\arraystretch}{0.94}
 \begin{tabular}{c|c|c|cccc|cc|c}
    \toprule[1pt]
\multirow{2}[4]{*}{Type} & \multirow{2}[4]{*}{Methods} & \multicolumn{1}{c|}{Scans Used} & \multicolumn{4}{c|}{Metrics} & \multicolumn{2}{c|}{{Inference Cost}} & \multirow{2}[4]{*}{\makecell[c]{{p-value↓}}} \\
\cmidrule{3-9}          &       & \multicolumn{1}{c|}{Label/Unlabel} & \multicolumn{1}{c}{DSC↑(\%)} & \multicolumn{1}{c}{Jaccard↑(\%)} & \multicolumn{1}{c}{95HD↓(voxel)} & \multicolumn{1}{c}{ASD↓(voxel)} & {Parameters} & {FLOPs} & \\
    \midrule
    Fully-supervised & UNet (Lower Bound) & 3/0   & 47.83 & 37.01 & 31.16 & 12.62 & {1.81M} & {3.00G} & {$\surd$} \\
    \midrule
    \multirow{8}[2]{*}{Semi-supervised} & UA-MT~\cite{yuUncertaintyAwareSelfensemblingModel2019} & 3/67  & 46.04 & 35.97 & 20.08 & 7.75 & {1.81M} & {3.00G} & {$\surd$} \\
          & SASSNet~\cite{liShapeAwareSemisupervised3D2020} & 3/67  & 57.77 & 46.14 & 20.05 & 6.06 & {1.81M} & {3.00G} & {$\surd$} \\
          & DTC~\cite{luoSemisupervisedMedicalImage2021}   & 3/67  & 56.90 & 45.67 & 23.36 & 7.39 & {1.81M} & {3.00G} & {$\surd$} \\
          & URPC~\cite{luoEfficientSemisupervisedGross2021} & 3/67  & 55.87 & 44.64 & 13.60 & 3.74 & {1.81M} & {3.00G} & {$\surd$} \\
          & MC-Net~\cite{wuSemisupervisedLeftAtrium2021}  & 3/67  & 62.85 & 52.29 & 7.62  & 2.33 & {1.81M} & {3.00G} & {$\surd$} \\
          & SS-Net~\cite{wuExploringSmoothnessClassSeparation2022} & 3/67  & 65.83 & 55.38 & 6.67  & 2.28 & {1.81M} & {3.00G} & {$\surd$} \\
          & BCP~\cite{baiBidirectionalCopyPasteSemiSupervised2023} & 3/67  & \underline{87.59} & \underline{78.67} & \underline{1.90}  & \underline{0.67} & {1.81M} & {3.00G} & {$\surd$} \\
          & \textbf{CVBM (Ours)} & \textbf{3/67}  & \textbf{87.85} & \textbf{79.03} & \textbf{1.82}  & \textbf{0.58} & {1.81M} & {3.00G} & {—} \\
    \midrule
    \midrule
    Fully-supervised & UNet (Lower Bound) & 7/0   & 79.41 & 68.11 & 9.35  & 2.70 & {1.81M} & {3.00G} & {$\surd$} \\
    \midrule
    \multirow{8}[2]{*}{Semi-supervised} & UA-MT~\cite{yuUncertaintyAwareSelfensemblingModel2019} & 7/63  & 81.65 & 70.64 & 6.88  & 2.02 & {1.81M} & {3.00G} & {$\surd$} \\
          & SASSNet~\cite{liShapeAwareSemisupervised3D2020} & 7/63  & 84.50 & 74.34 & 5.42  & 1.86 & {1.81M} & {3.00G} & {$\surd$} \\
          & DTC~\cite{luoSemisupervisedMedicalImage2021}   & 7/63  & 84.29 & 73.92 & 12.81 & 4.01 & {1.81M} & {3.00G} & {$\surd$} \\
          & URPC~\cite{luoEfficientSemisupervisedGross2021} & 7/63  & 83.10 & 72.41 & 4.84  & 1.53 & {1.81M} & {3.00G} & {$\surd$} \\
          & MC-Net~\cite{wuSemisupervisedLeftAtrium2021}  & 7/63  & 86.44 & 77.04 & 5.50  & 1.84 & {1.81M} & {3.00G} & {$\surd$} \\
          & SS-Net~\cite{wuExploringSmoothnessClassSeparation2022} & 7/63  & 86.78 & 77.67 & 6.07  & 1.40 & {1.81M} & {3.00G} & {$\surd$} \\
          & BCP~\cite{baiBidirectionalCopyPasteSemiSupervised2023}   & 7/63  & \underline{88.84} & \underline{80.62} & \underline{3.98}  & \underline{1.17} & {1.81M} & {3.00G} & {$\surd$} \\
          & \textbf{CVBM (Ours)} & \textbf{7/63}  & \textbf{89.98} & \textbf{82.30} & \textbf{1.37}  & \textbf{0.40} & {1.81M} & {3.00G} & {—} \\
    \midrule
    \midrule
     \multirow{4}[2]{*}{Fully-supervised} &  SAM~\cite{Kirillov_2023_ICCV} & -  & 59.39 & 44.60 & 9.63  & 3.07 & {93.74M} & {370.63G} & {$\surd$}  \\
    &SAM\_Med2D~\cite{chengSAMMed2D2023a} & -  & 76.13 & 64.54 & 5.06  & 1.24 & {271.24M} & {43.54G} & {$\surd$}  \\
    & UNet (Upper Bound) & 70/0  & 91.44 & 84.59 & 4.30  & 0.99 & {1.81M} & {3.00G} & {$\surd$} \\
    & {CVBM (Ours)} & {70/0}  & {91.79} & {85.11} & {1.01}  & {0.36} & {1.81M} & {3.00G} & {—} \\
    \bottomrule[1pt]
    \end{tabular}%
    }%
 \label{tab:ACDC}%
\end{table*}

\begin{figure*}[t]
\centering

\includegraphics[width=0.95\textwidth]{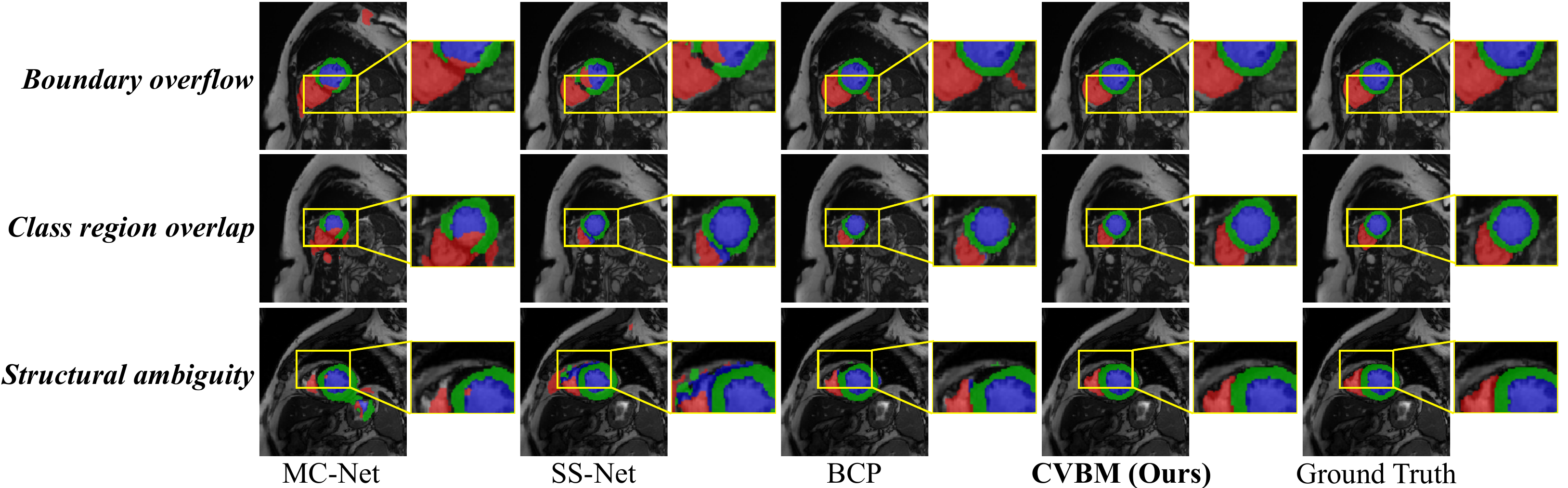} % 

\caption{Visualization results of ACDC dataset with 3/67 labeled data. CVBM reduces boundary overflow, region overlap, and structural ambiguity, leading to segmentation results across the three categories that more closely align with the ground truth. Best viewed by zoom-in on screen. }
\label{fig:ACDC_2D}

\end{figure*}

\subsubsection{ACDC Dataset}
In comparison to single-target segmentation, multi-target segmentation tasks present  more complex relationships between class boundaries. In these scenarios, the model must simultaneously model the feature relationships between classes and background, as well as among the different classes themselves. {Therefore, we further expanded our model to address multi-class segmentation tasks. Table \ref{tab:ACDC} illustrate the mean performance across three critical cardiac structures: myocardium, left ventricle, and right ventricle. {Under the setting of 3 labeled data, SS-Net employs pixel-level smoothness and inter-class separation, which promotes more compact foreground clustering, successfully improving the DSC from 47.83\% to 65.83\% {($p< 0.05$)}.} {BCP utilizes bidirectional copy-paste, reducing the distribution difference between labeled and unlabeled data, further increasing the DSC to 87.59\% {($p< 0.05$)}.} However, it is worth noting that as the number of labeled data increases (\ie, from 3 to 7), the performance of BCP decreases in terms of 95HD and ASD. In contrast, CVBM consistently exhibits improved performance in both the 3/67 and 7/63 labeled data configurations.} Remarkably, CVBM even surpasses the upper bound (70 labeled data) in terms of 95HD and ASD metrics. We also compared the results with SAM\_Med2D, and CVBM continued to achieve improved performance. {These findings demonstrated that CVBM is applicable to utilized to 2D multi-class segmentation, and with the number of labeled data increases, CVBM  effectively reduce uncertain regions at category boundaries, precisely delineating multiple organs.}

The visual representations of the ACDC dataset are illustrated in Fig. \ref{fig:ACDC_2D}. {In these visualizations, red denotes the right ventricle, green signifies the myocardium, and blue represents the left ventricle. In these exemplar cases, MC-Net, SS-Net, and BCP exhibit some discrete erroneous predictions in the right ventricle segmentation. In contrast, CVBM shows no discrete predictions, with clear shape and accurate structure. Furthermore, competing methods demonstrate phenomena, \eg, boundary overflow, class overlap, and structural ambiguity. Comparatively, the segmentation of CVBM exhibits more complete shapes, clearer category structures, and all class segmentation boundaries more closely approximate the ground truth. This demonstrates that under multi-class segmentation scenarios with complex category boundary relationships, CVBM  enhance the segmentation accuracy of boundaries, thereby producing  precise segmentation results.}

{In particular, under fully supervised training settings, CVBM consistently outperforms baseline methods across LA, Pancreas, and ACDC datasets on all metrics. On the LA dataset, CVBM achieves a 0.55\% improvement in DSC over VNet (92.02\% vs. 91.47\%) and more substantial gains in Jaccard index (+0.92\%), with similar improvements on the Pancreas and ACDC datasets. These results demonstrate the robust generalizability of the proposed cross-view modeling approach, highlighting its effectiveness for enhancing segmentation accuracy across diverse label proportions. Additionally, during inference, we only employ the foreground branch of the student model, maintaining consistency in parameter and FLOPs with competing methods across all three datasets.}
\begin{table}[htbp]

  \caption{{Comparisons with SOTA semi-supervised segmentation methods on HRF dataset.   Fully-supervise means training with all labeled samples on the V-Net backbone.}}

  \centering
  \scriptsize
  \resizebox{1.0\columnwidth}{!}{
    \renewcommand{\arraystretch}{0.98}
     \begin{tabular}{c|c|cccc}
    \toprule
    Labeled & Methods &\makecell[c]{DSC↑\\(\%)}  & \makecell[c]{Jaccard↑\\(\%)} &  \makecell[c]{95HD↓\\(voxel)} & \makecell[c]{ASD↓\\(voxel)} \\
    \midrule
    \multirow{8}[1]{*}{1/26} & UA-MT  \cite{yuUncertaintyAwareSelfensemblingModel2019} & 76.01 & 61.70 & 35.46 & 4.51 \\
          & SASSNet  \cite{liShapeAwareSemisupervised3D2020}  & 75.29 & 60.71 & 53.35 & 3.89 \\
          & DTC \cite{luoSemisupervisedMedicalImage2021}    & 76.02 & 61.62 & 35.12 & 3.62 \\
          & URPC  \cite{luoEfficientSemisupervisedGross2021} & 76.39 & 62.09 & 34.57 & 3.52 \\
          & MC-Net \cite{wuSemisupervisedLeftAtrium2021} & 77.01 & 62.89 & 32.61 & 3.73 \\
          & SS-Net \cite{wuExploringSmoothnessClassSeparation2022} & 77.50 & 63.51 & 28.78 & 4.23 \\
          & BCP  \cite{baiBidirectionalCopyPasteSemiSupervised2023}  & 77.14 & 63.05 & 24.15 & 4.50 \\
          & \textbf{CVBM (Ours)  } & \textbf{78.73} & \textbf{65.07} & \textbf{20.89 }& \textbf{3.21} \\
    \midrule
    \multirow{8}[2]{*}{3/24} & UA-MT \cite{yuUncertaintyAwareSelfensemblingModel2019}  & 76.58 & 62.41 & 28.82 & 4.08 \\
          & SASSNet \cite{liShapeAwareSemisupervised3D2020} & 77.58 & 63.70 & 29.83 & 3.52 \\
          & DTC  \cite{luoSemisupervisedMedicalImage2021} & 76.80 & 62.71 & 28.85 & 3.15 \\
          & URPC \cite{luoEfficientSemisupervisedGross2021}   & 77.08 & 63.08 & 27.73 & 3.14 \\
          & MC-Net  \cite{wuSemisupervisedLeftAtrium2021} & 77.11 & 63.09 & 27.39 & 3.39 \\
          & SS-Net  \cite{wuExploringSmoothnessClassSeparation2022} & 78.07 & 64.31 & 25.67 & 3.26 \\
          & BCP  \cite{baiBidirectionalCopyPasteSemiSupervised2023}   & 78.77 & 65.19 & 21.04 & 3.43\\
          & \textbf{CVBM (Ours)}  & \textbf{79.11} & \textbf{65.69} & \textbf{20.80} & \textbf{3.12} \\
    \midrule
    {27/0} & {Fully-supervised} & {80.36} & {67.38} & {23.96} & {2.40} \\
    \bottomrule
    \end{tabular}}

  \label{tab:hrf}%
\end{table}%
\begin{figure*}[t]
\centering
{\includegraphics[width=0.95\textwidth]{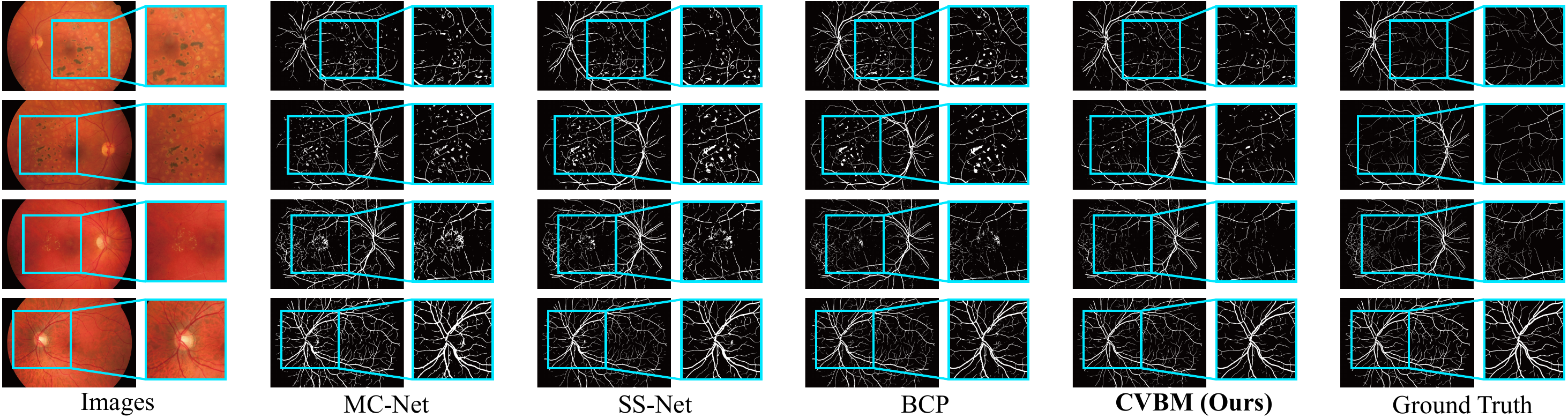}}

\caption{{Visualization results of  HRF dataset with 1 labeled data. GT means Ground Truth. Best  viewed by zoom-in on screen.}  }

\label{fig:vessel3}
\end{figure*}
{
\subsubsection{HRF Dataset}
The HRF dataset has an RGB three-channel format similar to natural images. However, its vascular network is far more complex topologically, and retinal vessels have much lower chromatic contrast with their backgrounds compared to natural scenes, making segmentation particularly challenging.
Table \ref{tab:hrf} presented the comparison between CVBM and SOTA methods on the HRF dataset. Notably, CVBM outperformed existing SOTA methods in the setting of both 1/26 and 3/24 labeled samples. Furthermore, as the number of labeled samples decreased, the advantage of CVBM became more pronounced. For instance, in terms of 95HD, CVBM surpassed BCP by 0.24 under the 3/24 labeled setting and by 3.26 under the 1/26 labeled setting. This improvement demonstrated that even in the challenging task of vessel segmentation with complex boundaries and diverse shapes, CVBM accurately identified and located vessel regions.

Figure \ref{fig:vessel3} presented the results of retinal vessel segmentation. It was evident that the proposed CVBM effectively segmented relatively continuous vessels even in the presence of lesions, as shown in the top three lines. Notably, our method improved the identification of small vessels at arterial and venous endpoints, especially at their intersections, as illustrated in the last line. This further indicates that, in regions with high prediction uncertainty (\eg, lesions and small vessels), our method assisted the foreground model reduce ambiguous predictions and improve confidence.

Above experiments confirm that CVBM not only achieves superior performance in 3D medical imaging but also maintains strong efficacy in 2D image segmentation. Crucially, the method effectively handles challenging scenarios characterized by ambiguous tissue boundaries and low contrast between foreground/background structures, thereby validating its applicability across 2D natural imaging modalities.}

\subsection{Ablation Study} \label{AB}
To enhance comprehension of CVBM, in this subsection, we conduct extensive ablation experiments to evaluate the impact of each component in the model on semi-supervised medical image segmentation performance.

\begin{table}[htbp]

 \caption{{Ablation of different components on LA and ACDC datasets with 10\% labeled data. $\mathcal{T}_\text{fg}$ and $\mathcal{T}_\text{bg}$ denotes the foreground and background modeling task, respectively. $Mix$ represents the mixing layer.} }

 \centering
 \large
 \resizebox{1.0\columnwidth}{!}{
 \renewcommand{\arraystretch}{0.98}
 \begin{tabular}{c|c|cccc|cccc}
 \toprule[1pt]
 Dataset & Methods & $\mathcal{T}_\text{fg}$ & $\mathcal{T}_\text{bg}$ & $Mix$ & $\mathcal{L}_\text{bcl}$ &\makecell[c]{DSC↑\\(\%)} & \makecell[c]{Jaccard↑\\(\%)} & \makecell[c]{95HD↓\\(voxel)} & \makecell[c]{ASD↓\\(voxel)} \\
 \midrule
 \multirow{5}{*}{LA} & Supervised & & & & & 82.74 & 71.72 & 13.35 & 3.26 \\
 \cmidrule{2-10}
 & \#1 & $\surd$ & & & & 87.14 & 77.47 & 9.15 & 3.04 \\
 & \#2 & $\surd$ & $\surd$ & & & 89.62 & 81.33 & 7.81 & 1.76 \\
 & \#3 & $\surd$ & $\surd$ & $\surd$ & & 90.23 & 82.30 & 6.47 & 1.72 \\
 \cmidrule{2-10}
 & {\#4} & {$\surd$} & {$\surd$} & {$\surd$} & {$\surd$} & {\textbf{91.19}} & {\textbf{83.87}} & {\textbf{5.45}} & {\textbf{1.16}} \\
 \midrule
 \multirow{5}{*}{{ACDC}} & {Supervised} & {} & {} & {} & {} & {79.41} & {68.11} & {9.35} & {2.70} \\
 \cmidrule{2-10}
 & {\#1} & {$\surd$} & {} & {} & {} & {81.04} & {69.95} & {6.02} & {2.21} \\
 & {\#2} & {$\surd$} & {$\surd$} & {} & {} & {84.23} & {73.62} & {3.72} & {1.01} \\
 & {\#3} & {$\surd$} & {$\surd$} & {$\surd$} & {} & {87.12} & {79.45} & {3.13} & {0.84} \\
 \cmidrule{2-10}
 & {\#4} & {$\surd$} & {$\surd$} & {$\surd$} & {$\surd$} & {\textbf{89.98}} & {\textbf{82.30}} & {\textbf{1.37}} & {\textbf{0.40}} \\
 \bottomrule[1pt]
 \end{tabular}}%
 \label{tab:ablation}%

\end{table}%
 
\begin{figure*}[t]
\centering
{\includegraphics[width=0.95\textwidth]{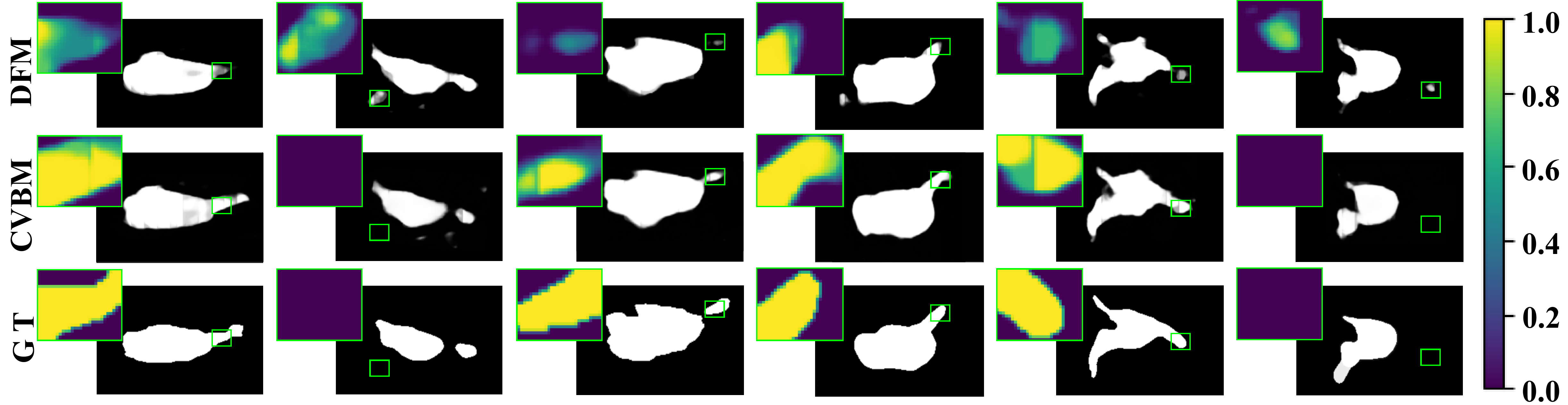}}

\caption{{Confidence maps comparison between traditional foreground-oriented modeling (DFM) and our CVBM. GT indicates Ground Truth. Solid yellow and purple represent high-confidence foreground and background predictions, respectively.}}

\label{fig:confidence_map}
\end{figure*}

\begin{figure}[t]
\centering
{\includegraphics[width=0.95\columnwidth]{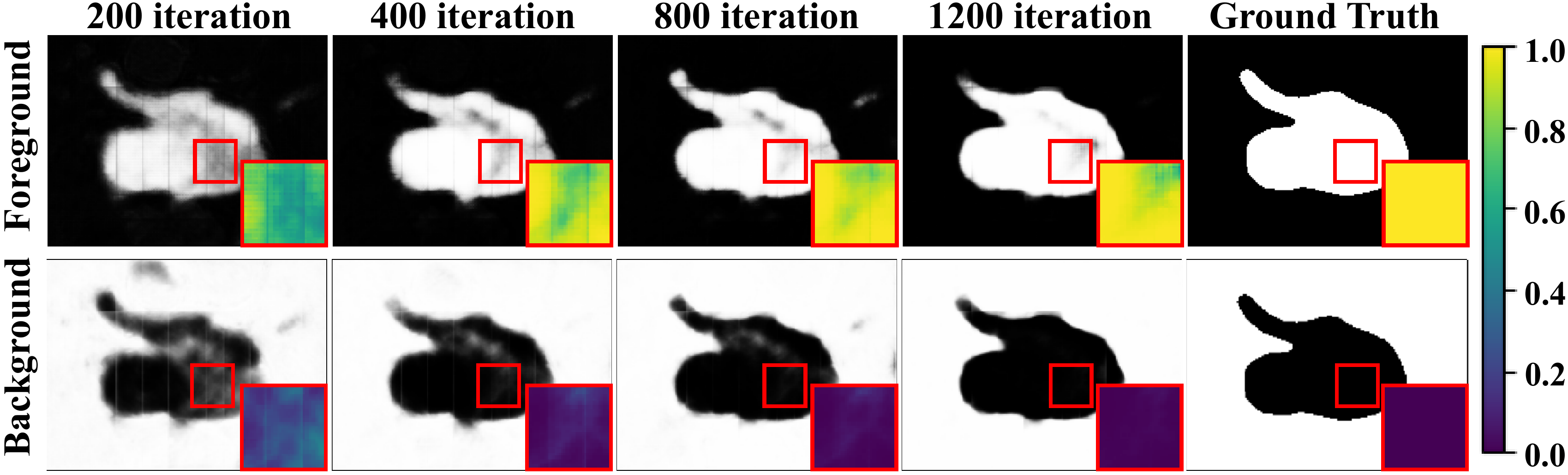}}

\caption{{The impact of  high-confidence backgrounds on foreground segmentation. The red box shows the confidence maps where yellow indicates higher foreground confidence and purple indicates higher background confidence.}}
\label{fig:fg_bg_map}

\end{figure}

\subsubsection{Effectiveness of different components of CVBM} 
{The ablation of each component within CVBM is illustrated in Table~\ref{tab:ablation}. When compared to the supervised baseline, our method enhances segmentation performance, yielding improvements in DSC (91.19\% vs. 82.74\%) and Jaccard scores (83.87\% vs. 71.72\%). Our method shows similar improvements on the ACDC dataset, achieving a DSC of 89.98\% (vs. 79.41\%) and a Jaccard of 82.30\% (vs. 68.11\%) compared to the supervised baseline. A comparative analysis of \#1 and \#2 reveals that bidirectional cross-view modeling (\#2) outperforms single-perspective modeling (\#1) across both datasets. On ACDC, this pattern is also evident, with bidirectional modeling (\#2) showing DSC of 84.23\% compared to single-perspective modeling's (\#1) 81.04\%. Additionally, we isolated the mixing layer and $\mathcal{L}_\text{bcl}$ to verify their contributions to CVBM (\ie, \#3 and \#4). Evidently, the connection between foreground predictions and background-guided predictions through the mixing layer (\#3) improve performance to 90.23\% on LA and 87.12\% on ACDC. And the enhancement of \#4 further prove the effectiveness of the bidirectional consistency optimization strategy in guiding the model to refine foreground predictions, with final DSC scores reaching 91.19\% and 89.98\% on LA and ACDC respectively. The above ablation experiments demonstrate that each component within CVBM can facilitate the foreground model in producing accurate segmentation results. }
{\subsubsection{Impact on foreground modeling}
Figure~\ref{fig:fg_bg_map} provides a comparative visualization of segmentation outcomes across multiple tasks as iteration count increases (200, 400, 800, 1200).  Within these regions, background predictions exhibit more uniform purple coloration compared to foreground predictions, indicating higher confidence in the background modeling. As iterations increase, green areas in foreground predictions progressively diminish, demonstrating enhanced foreground prediction confidence. This observation substantiates the theoretical proof in Theorem 2, which demonstrates that high-confidence background modeling effectively enhances foreground predictive confidence.}
\begin{table}[htbp]

\caption{Ablation of the model structure on LA dataset with 4 and 8 labeled data.}

 \centering
 \small
 \resizebox{1.0\columnwidth}{!}{
 \renewcommand{\arraystretch}{0.98}
 \begin{tabular}{c|l|cccc}
 \toprule[1pt]
 Label & Methods &\makecell[c]{DSC↑\\(\%)} & \makecell[c]{Jaccard↑\\(\%)} & \makecell[c]{95HD↓\\(voxel)} & \makecell[c]{ASD↓\\(voxel)} \\
 \midrule
 & Dual foreground modeling & 88.45 & 79.41 & 7.90 & 2.11 \\
 4/76 & Dual background modeling & 88.61 & 79.65 & 7.16 & 2.12 \\
 & {\textbf{CVBM (Ours)}} & {\textbf{89.50}} & {\textbf{81.07}} &{\textbf{5.78}} & {\textbf{2.10}} \\
 \midrule
 & Dual foreground modeling & 90.03 & 81.96 & 6.63 & 1.68 \\
 8/72 & Dual background modeling & 90.06 & 82.05 & 6.86 & 1.74 \\
 & {\textbf{CVBM (Ours)}} & {\textbf{91.19}} & {\textbf{83.87}} & {\textbf{5.45}} & {\textbf{1.61}} \\
 \bottomrule[1pt]
 \end{tabular}}%
 \label{tab:model}%

\end{table}%

\subsubsection{Advantages of cross-view modeling} To evaluate the efficacy of the cross-view modeling structure, we designed and compared two variant models: Dual Foreground Modeling (DFM) and Dual Background Modeling (DBM). In the latter case, where the primary segmentation target is the background, we extract regions with prediction scores below 0.5 (\ie, foreground regions). The comparative results are presented in Table \ref{tab:model}. Notably, while both single-view models achieved good segmentation performance, the cross-view modeling structure (\ie, CVBM) consistently outperformed these variants across all four evaluation metrics under both the 4/76 and 8/72 labeled data configurations. These results indicate that compared to cross-view modeling, single-view modeling shows reduced prediction confidence. In contrast, CVBM leverages background modeling to acquire cross-view features and iteratively refines the discrepancies between foreground predictions and background-guided predictions by $L_\text{bcl}$, enabling the teacher model to produce more reliable pseudo-labels. {Furthermore, as demonstrated in Fig.~\ref{fig:confidence_map}, cross-view modeling significantly reduces prediction uncertainty in boundary pixels compared to traditional foreground-oriented modeling approaches, thereby producing reliable predictions. The improved performance of CVBM underscores the efficiency of integrating multiple perspectives in tackling complex medical image segmentation tasks, particularly in scenarios with limited labeled data.}

\begin{table}[htbp]

\caption{Ablation of bidirectional consistency on LA and ACDC dataset with 4 and 3 labeled data, respectively.}

 \centering
 \small
 \resizebox{1.0\columnwidth}{!}{
 \renewcommand{\arraystretch}{1.1}
 \begin{tabular}{c|l|cccc}
 \toprule[1pt]
 Dataset & Methods &\makecell[c]{DSC↑\\(\%)} & \makecell[c]{Jaccard↑\\(\%)} & \makecell[c]{95HD↓\\(voxel)} & \makecell[c]{ASD↓\\(voxel)} \\
 \midrule
 \multirow{4}[4]{*}{LA} & {Baseline} & 87.12 & 77.44 & 10.11 & 2.65 \\
\cmidrule{2-6} & +Direct Consistency & 89.24 & 80.65 & 7.68 & 2.55 \\
 & +Inverse Consistency & 88.97 & 80.22 & 7.96 & 2.52 \\
 & {\textbf{+Bidirectional Consistency}} & {\textbf{89.50}} & {\textbf{81.07}} & {\textbf{5.78}} & {\textbf{2.10}} \\
 \midrule
 \multirow{4}[4]{*}{ACDC} & Baseline & 85.73 & 75.79 & 4.78 & 1.34 \\
\cmidrule{2-6} & +Direct Consistency & 86.86 & 77.52 & 2.66 & 0.69 \\
 & +Inverse Consistency & 86.03 & 76.28 & 4.15 & 1.20 \\
 & {\textbf{+Bidirectional Consistency}} & {\textbf{87.85}} & {\textbf{79.03}} & {\textbf{1.80}} & {\textbf{0.58}} \\
 \bottomrule[1pt]
 \end{tabular}}%

 \label{tab:loss}%
\end{table}%

\subsubsection{Ablation of bidirectional consistency loss} 
To investigate the impact of Bidirectional Consistency Loss on model performance, we decompose it into two components: Direct Consistency Loss and Inverse Consistency Loss. The baseline is CVBM without the bidirectional consistency. Results are presented in Table \ref{tab:loss}. Apparently, employing Direct Consistency Loss to align foreground predictions exhibits better performance than the baseline. Surprisingly, the addition of Inverse Consistency Loss also demonstrates performance improvements. {This indicates that utilizing background modeling prediction to constraint foreground model learning is feasible and beneficial.} Moreover, integrating both consistency losses can further achieve the best performance. {The results indicate that our proposed bilateral contrastive loss ($\mathcal{L}_\text{bcl}$), which encourages the foreground model to achieve consistent segmentation across multiple viewpoints, {enhances the predictive accuracy of foreground modeling.}}

\begin{table}[htbp]

  \centering
  \small
  \caption{Performance comparison of different output strategies of CVBM across LA, Pancreas, and ACDC datasets.}

  \resizebox{1.0\columnwidth}{!}{
   \renewcommand{\arraystretch}{1.1}
  \begin{tabular}{c|l|cccc}
    \toprule[0.8pt]
    Dataset  & Output &\makecell[c]{DSC↑\\(\%)} & \makecell[c]{Jaccard↑\\(\%)} & \makecell[c]{95HD↓\\(voxel)} & \makecell[c]{ASD↓\\(voxel)} \\
    \midrule
    \multirow{2}[2]{*}{LA}  
    & Foreground result (CVBM)
    & 89.50 & 81.07 & 5.78  & 2.10 \\
    & Mix layer result (CVBM+)  & \underline{90.12} & \underline{81.65} & \underline{5.04}  & \underline{1.92} \\
    & \textbf{Average result (CVBM++)} & \textbf{90.47} & \textbf{82.69} & \textbf{4.94}  & \textbf{1.73} \\
    \midrule
    \multirow{2}[2]{*}{Pancreas}  
    & Foreground result (CVBM) & 83.65 & 72.16 & 4.48  & 1.30 \\
    & Mix layer result (CVBM+)  & \underline{84.02} & \underline{73.81} & \underline{3.63}  & \underline{1.16} \\
    & \textbf{Average result (CVBM++)} & \textbf{84.61} & \textbf{74.01} & \textbf{3.54}  & \textbf{1.01} \\
    \midrule
    \multirow{2}[2]{*}{ACDC}  
    & Foreground result (CVBM) & 87.85 & 79.03 & 1.82  & 0.58 \\
    & Mix layer result (CVBM+)  & \underline{87.98} & \underline{79.94} & \underline{1.60}  & \underline{0.53} \\
    & \textbf{Average result (CVBM++)} & \textbf{88.24} & \textbf{80.45} & \textbf{1.46}  & \textbf{0.50} \\
    \bottomrule[0.8pt]
  \end{tabular}
  }

\label{tab:output strategies}
\end{table}

{
\subsubsection{Quantitative assessment of alternative output configurations} We explored two alternative output configurations for CVBM: the average result $((Q_\text{fg}+Q_\text{bg})/2)$ and {the mix layer result $(Q_M)$}. As shown in Table~\ref{tab:output strategies}, both configurations consistently outperformed the foreground branch across all evaluation metrics and datasets.  Specifically, the average result achieves the highest performance improvements, with DSC increases of 0.97\%, 0.96\%, and 0.39\% on the LA, Pancreas, and ACDC datasets, respectively. These findings demonstrate that background modeling improves the predictive confidence of foreground modeling, as combining background and foreground predictions leads to highest segmentation accuracy. Although these alternative strategies yield performance gains, we opted to report only the foreground branch results  to ensure fair comparison with existing sota methods in terms of computational complexity and model size.

\begin{figure}[htbp]

\centering
{\includegraphics[width=0.95\columnwidth]{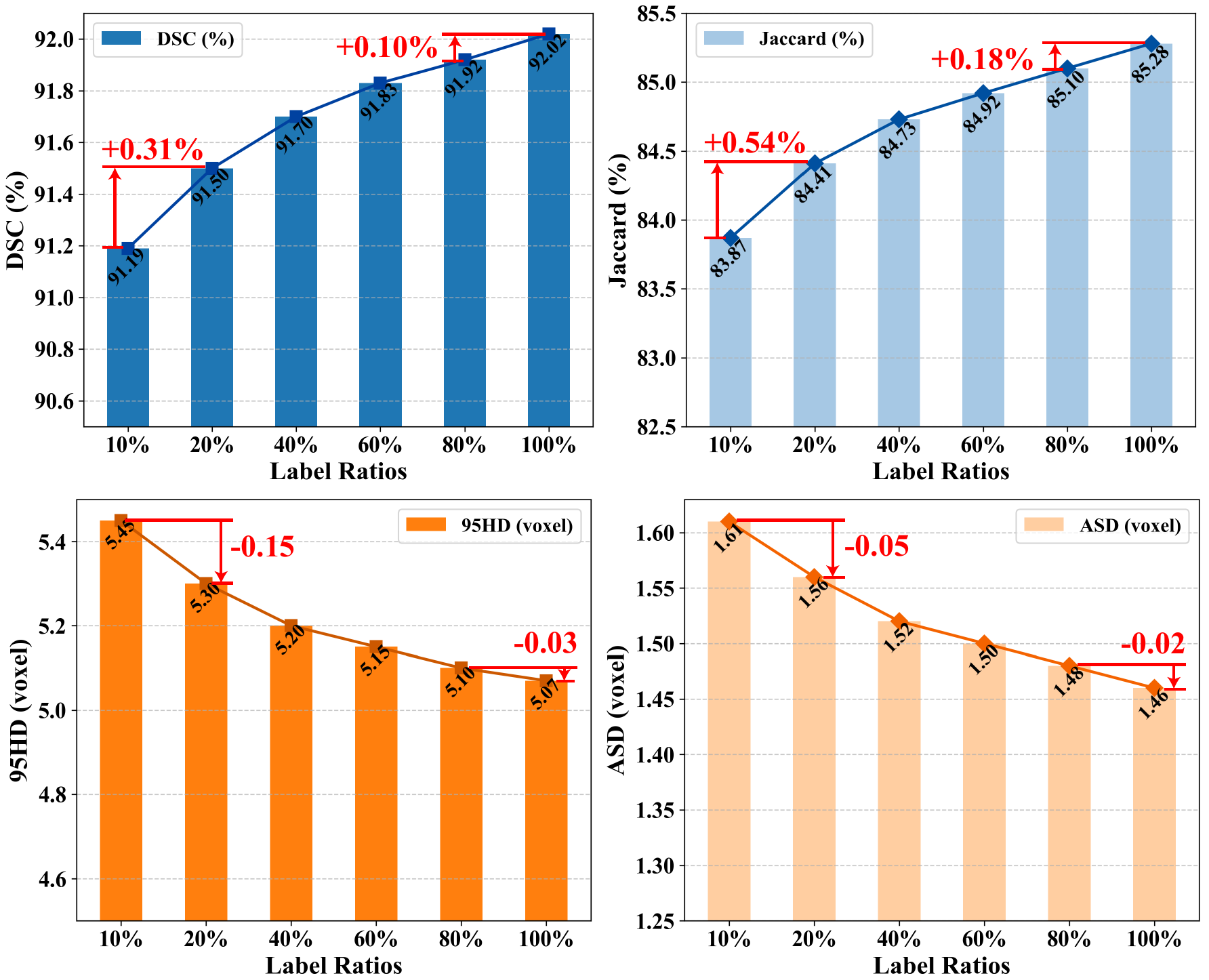}}

\caption{Label Ratio Impact on Model Performance. Experiments conducted utilizing the LA dataset.}
\label{fig:chart}

\end{figure}
\subsubsection{Impact of labeling ratios} In Fig.~\ref{fig:chart}, our method shows consistent performance improvement as labeled data increases from 10\% to 100\%. DSC rises from 91.19\% to 92.02\% (+0.83\%), while Jaccard increases from 83.87\% to 85.28\% (+1.41\%).
Error metrics decrease, confirming better segmentation accuracy: 95HD reduces from 5.43 to 5.07 voxels (-6.63\%), and ASD decreases from 1.61 to 1.46 voxels (-9.32\%).
Performance gains are most significant at lower labeling rates (10\%→40\%) and plateau at higher rates (60\%→100\%). From 10\% to 60\%, DSC increases by 0.62\%, but only by 0.21\% from 60\% to 100\%. Similarly, Jaccard improves by 1.05\% in the first interval but only 0.36\% in the second. This non-linear pattern confirms our method's effectiveness, particularly in enhancing foreground modeling prediction confidence under low labeling conditions.

\begin{table}[htbp]

\caption{Ablation study of different foreground ratios on the LA dataset with 4 labeled data.}

 \centering
  \renewcommand{\arraystretch}{1.0}
 \begin{tabular}{c|cccc}
 \toprule
  Ratios & DSC↑(\%) & Jaccard↑(\%) & 95HD↓(voxel) & ASD↓(voxel)\\
 \midrule
  0.5 & 80.98 & 69.84 & 10.29 & 4.18\\
  0.8 & 86.56 & 76.30 & 7.70& 3.28 \\
 1.0 & {{89.50}} & {{81.07}} &{{5.78}} & {{2.10}} \\
 1.2 & 90.07 & 82.04 & 5.44 & 1.83 \\
 1.5 & \textbf{91.98} & \textbf{83.69} & \textbf{4.92} & \textbf{1.52} \\
 \bottomrule
 \end{tabular}
 \label{tab:foreground-to-background ratios}%

\end{table}
\subsubsection{Impact of different foreground ratios}
To analyze the effect of varying foreground proportions, we extracted foreground regions using ground truth masks, applied scaling factors (0.5-1.5), and reintegrated these modified foregrounds into the original images. Table~\ref{tab:foreground-to-background ratios} illustrates performance across different scaling ratios. As the ratio increased from 0.5 to 1.5, we observed substantial improvements: DSC increased by 11.0 percentage points (80.98\% to 91.98\%) while 95HD decreased by 52.2\% (10.29 to 4.92). These findings clearly demonstrate that higher foreground ratios correlate with enhanced performance. This phenomenon aligns with Theorem 2, which establishes that CVBM strengthens prediction confidence for single-voxel foreground modeling through background-assisted modeling. Increased foreground ratios expand the boundary interface, generating more uncertain voxels for optimization. Consequently, a greater number of foreground pixels contribute to entropy reduction, yielding improved overall performance.}

\begin{figure}[htbp]

\centering
\includegraphics[width=0.95\columnwidth]{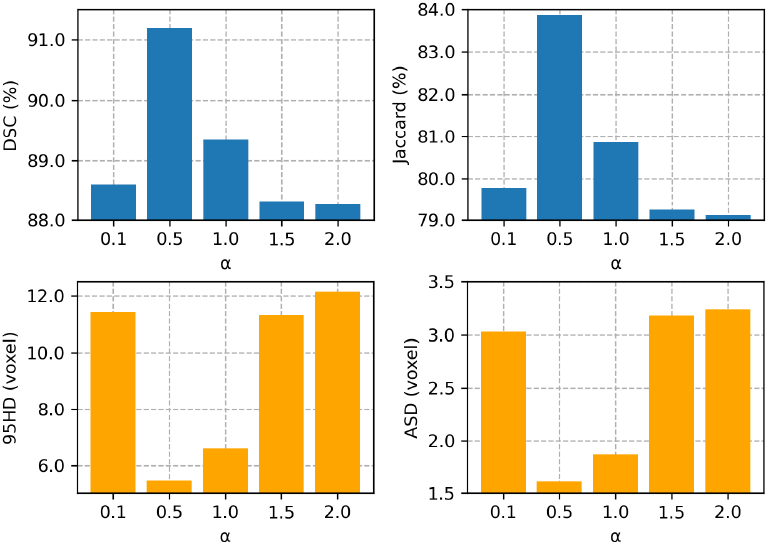} % Reduce the figure size so that it is slightly narrower than the column. Don't utilize precise values for figure width.This setup will avoid overfull boxes.

\caption{Variations of four evaluation metrics at different values of parameter $\alpha$. {utilizing}  10\% labeled data on LA dataset. \textcolor{myblue}{\textbf{Blue}} indicates higher performance is better, while \textcolor{myyellow}{\textbf{yellow}} indicates that lower performance is better. }

\label{fig:BAR}
\end{figure}
\subsubsection{Quantitative analysis on hyper-parameter}
Fig. \ref{fig:BAR} illustrates the variation of four evaluation metrics when adjusting the weighting factor $\alpha$ within the range of 0.1 to 2.0. Notably, the model attains optimal performance when $\alpha$ is set to 0.5. Both decreasing and increasing $\alpha$ result in performance degradation. At lower $\alpha$ values, the model underutilizes the potential of unlabeled samples, limiting model ability to learn  anatomical information from unlabeled data. Conversely, higher $\alpha$ values amplify the influence of inaccurate pseudo-labels, introducing undesirable artifacts into the learning process. The optimal performance observed at $\alpha=0.5$ suggests an ideal balance between utilizing unlabeled data information and maintaining robustness against potential noise. Based on this analysis, we adopt $\alpha=0.5$ as the standard setting throughout this study, balancing the utilization of unlabeled data with model reliability.

\section{Conclusion}

In this study, we breaks the trend of recent SOTAs that predominantly prioritizing foreground modeling, introducing a novel Cross-view Bidirectional Modeling scheme (CVBM) for semi-supervised medical image segmentation. CVBM integrates background modeling to enhance foreground model predictive confidence. Additionally, we propose a bidirectional consistency optimization scheme that encourages bidirectional alignment between foreground predictions and background predictions. Extensive experiments demonstrated that CVBM obtains new SOTA performance on popular SSMIS benchmarks. For example, CVBM achieves a higher DSC of 84.57\% on Pancreas dataset {utilizing}  only 12 labeled volumes compared to the fully supervised baseline of 83.89\% {utilizing}  62 labeled volumes. Additionally, we ensure that no additional computational overhead during our inference phase. {Future research will explore various cross-view modeling scenarios. In unsupervised learning, the complementary relationship between foreground and background serves as a self-supervised signal, providing bidirectional optimization space constraints. In active learning, sample selection can be guided through inconsistencies between foreground and background predictions, prioritizing the annotation of samples that provide the most value for model improvement.}

\bibliographystyle{ieeetr}
\bibliography{IEEEabrv,mainbib}
\clearpage

\section{Appendix}
\subsection{Symbols and Definitions}
To facilitate clarity and consistency throughout the paper, we present a comprehensive compilation of all symbols and notations used in this work. Clear mathematical notation is essential for understanding our proposed contrastive volumetric background modeling (CVBM) framework. Table \ref{tab:symbols} provides a systematic overview of all mathematical symbols and their definitions. This reference table enables readers to quickly look up any unfamiliar notation, understand the precise meaning of each symbol in context, follow the mathematical derivations with greater ease, and avoid potential confusion from ambiguous notation.

\begin{table*}[htbp]
\centering
\caption{Summary of Symbols and Definitions}

\resizebox{1.0\textwidth}{!}{
\renewcommand{\arraystretch}{1.0}
\begin{tabular}{llllll}
\toprule
\textbf{Symbol} & \textbf{Definition} & \textbf{Symbol} & \textbf{Definition} & \textbf{Symbol} & \textbf{Definition} \\
\midrule
\( Y \) & Binary tensor & \( Y_\text{bg} \) & Background label & \( Y_\text{M} \) & Multi-class ground truth \\
\( Y_\text{M,bg} \) & Multi-target background label & \( X^{a}, X^{b} \) & Augmented labeled data & \( \mathcal{M} \) & Binary mask \\
\( \beta \) & Scaling factor & \( Q_\text{fg}, Q_\text{bg} \) & Foreground/background predictions & \( Q_\text{M} \) & Mixed prediction \\
\( \mathcal{L}_\text{fg}, \mathcal{L}_\text{bg} \) & Foreground/background loss & \( \mathcal{L}_\text{M} \) & Mixed loss & \( \mathcal{L}_\text{rw} \) & Region-wide loss \\
\( \mathcal{L}_\text{bcl} \) & Bidirectional consistency loss & \( \mathcal{L}_\text{total} \) & Total loss & \( \mathcal{E} \) & Shared encoder \\
\( \mathcal{D}_\text{fg}, \mathcal{D}_\text{bg} \) & Foreground/background decoders & \( \psi \) & Convolution operation & \( H_A, H_B \) & Architecture entropy \\
\( \epsilon_1, \epsilon_2 \) & Consistency constraints & \( \alpha, \lambda \) & Loss balancing factors & \( P_\text{fg}, P_\text{bg} \) & Pseudo-labels \\
\( \hat{Y}_\text{fg}, \hat{Y}_\text{bg} \) & Augmented labels & \( \hat{X}^{a}, \hat{X}^{b} \) & Augmented inputs & \( \text{onehot}(\cdot) \) & One-hot encoding \\
\( \odot \) & Element-wise multiplication & \( \text{concat}(\cdot) \) & Concatenation operation & \( \mathcal{L}_\text{seg} \) & Segmentation loss \\
\( \mathcal{L}_\text{mse} \) & Mean Squared Error loss & \( D_{\text{fg1}}, D_{\text{fg2}} \) & Dual foreground decoders & \( D_{\text{fg}}, D_{\text{bg}} \) & Foreground/background decoders \\
\( H(\mu) \) & Binary entropy & \( \mu \) & Foreground prediction probability & \( q \) & Background prediction probability \\
\bottomrule
\end{tabular}}
\label{tab:symbols}

\end{table*}

{
\subsection{Theoretical Analysis}\label{sec: Theoretical Analysis}
In this subsection, we provide theoretical insights into background-assisted modeling, demonstrating that background-assisted training paradigm reduce prediction uncertainty compared to traditional foreground-oriented training methods.

\noindent \textbf{Notations:}
Given an input image \( X \) and output probability \( Y \), a shared encoder \( \mathcal{E}: X \to H \) maps \( X \) to a latent space \( H \). \textbf{Architecture A} (foreground-oriented approach) employs two foreground decoders \( D_{\text{fg1}} \) and \( D_{\text{fg2}} \), where \( D_{\text{fg1}}(h) \) and \( D_{\text{fg2}}(h) \) estimate the probability of pixel \( p \) belonging to the foreground. \textbf{Architecture B} (background-assisted model) uses a foreground decoder \( D_{\text{fg}} \) and a background decoder \( D_{\text{bg}} \), with \( D_{\text{fg}}(h) \) and \( D_{\text{bg}}(h) \) estimating the probabilities of \( p \) belonging to the foreground and background, respectively. The uncertainty for each decoder is defined as  $H_{A1}(h) = -[D_{\text{fg1}}(h) \log(D_{\text{fg1}}(h)) + (1-D_{\text{fg1}}(h)) \log(1-D_{\text{fg1}}(h))]$ and $H_{A2}(h) = -[D_{\text{fg2}}(h) \log(D_{\text{fg2}}(h)) + (1-D_{\text{fg2}}(h)) \log(1-D_{\text{fg2}}(h))]$ for Architecture A, and $H_{B1}(h) = -[D_{\text{fg}}(h) \log(D_{\text{fg}}(h)) + (1-D_{\text{fg}}(h)) \log(1-D_{\text{fg}}(h))]$ and $H_{B2}(h) = -[D_{\text{bg}}(h) \log(D_{\text{bg}}(h)) + (1-D_{\text{bg}}(h)) \log(1-D_{\text{bg}}(h))]$ for Architecture B. The architecture entropy is then \( H_A(p) = H_{A1}(p) + H_{A2}(p) \) for Architecture A and \( H_B(p) = H_{B1}(p) + H_{B2}(p) \) for Architecture B.

\noindent \textbf{Lemma 1 (Lower Bound of Uncertainty for Architecture A)}
Under the consistency constraint\( ||D_\text{fg1}(h) - D_\text{fg2}(h)||^2 \leq \epsilon_1 \), for any pixel \( p \) belonging to the foreground:
\begin{equation}\label{eq:Lower_A}
H_A(p) \geq -2[\mu \log(\mu) + (1-\mu) \log(1-\mu)]- \sqrt{\epsilon_1} \log(\sqrt{\epsilon_1}),
\end{equation}
where \( \mu = D_{\text{fg1}}(h) \).

\noindent \textbf{Proof:} The consistency constraint ensures that the two foreground decoders produce similar outputs: \( |D_{\text{fg1}}(h) - D_{\text{fg2}}(h)| \leq \sqrt{\epsilon_1} \). 
To minimize \( H_A(p) \), we consider the case where \( D_{\text{fg2}}(h) = \mu + \sqrt{\epsilon_1} \). Substituting this into the entropy expression and using a Taylor expansion, along with simple algebraic transformations, for small \( \sqrt{\epsilon_1} \), we obtain:
\begin{equation}
\begin{split}
H_A(p) 
&\approx-2[\mu \log(\mu) + (1-\mu) \log(1-\mu)]\\
&- \sqrt{\epsilon_1}\log(\sqrt{\epsilon_1}) - \sqrt{\epsilon_1}\log(\frac{\mu}{\sqrt{\epsilon_1}})\\
&+\sqrt{\epsilon_1}\log(1-\sqrt{\epsilon_1}) + \sqrt{\epsilon_1}\log(\frac{1 - \mu}{1-\sqrt{\epsilon_1}}).\\
\end{split}
\end{equation}

For small~$\sqrt{\epsilon_1}$, the terms $\sqrt{\epsilon_1}\log(\frac{\mu}{\sqrt{\epsilon_1}})$, 
$\sqrt{\epsilon_1}\log(1-\sqrt{\epsilon_1})$,  $\sqrt{\epsilon_1}\log(\frac{1 - \mu}{1-\sqrt{\epsilon_1}})$  become negligible compared to $\sqrt{\epsilon_1}\log(\sqrt{\epsilon_1})$, which larger in magnitude.
Therefore, under the consistency constraint, for any pixel $p$:
\begin{equation}\label{eq:entropy_gradient}
H_A(p) \geq -2[\mu \log(\mu) + (1-\mu) \log(1-\mu)]- \sqrt{\epsilon_1} \log(\sqrt{\epsilon_1}),
\end{equation}
where $\mu = D_{\text{fg1}}(h)$. \hfill \(\square\)

\noindent \textbf{Lemma 2 (Upper Bound of Uncertainty for Architecture B)}
Under the inverse consistency constraint\( ||D_\text{fg}(h) + D_\text{bg}(h) - 1||^2 \leq \epsilon_2 \), for any pixel \( p \):
\begin{equation}\label{eq:upper_b}
H_B(p) \leq -2[\mu \log(\mu) + (1-\mu) \log(1-\mu)]+ \sqrt{\epsilon_2} \log(\sqrt{\epsilon_2}),
\end{equation}
where \( \mu = D_{\text{fg}}(h) \).

\noindent \textbf{Proof:} The inverse consistency constraint implies \( |D_{\text{fg}}(h) + D_{\text{bg}}(h) - 1| \leq \sqrt{\epsilon_2} \). Let \( \mu = D_{\text{fg}}(h) \), so \( D_{\text{bg}}(h) = 1 - \mu \pm \delta \), where \( \delta \leq \sqrt{\epsilon_2} \). 
To maximize \( H_B(p) \), we consider the case where \( D_{\text{bg}}(h) = 1 - \mu + \sqrt{\epsilon_2} \). For small \( \sqrt{\epsilon_2} \), we use a Taylor expansion, along with simple algebraic transformations:
\begin{equation}
\begin{split}
H_B(p)
&\approx-2[\mu \log(\mu) + (1-\mu) \log(1-\mu)]\\
&-\sqrt{\epsilon_2}\log(1-\sqrt{\epsilon_2}) - \sqrt{\epsilon_2}\log(\frac{1 - \mu}{1-\sqrt{\epsilon_2}}) \\
&+ \sqrt{\epsilon_2}\log(\sqrt{\epsilon_2}) + \sqrt{\epsilon_2}\log(\frac{\mu}{\sqrt{\epsilon_2}})
\end{split}
\end{equation}
For small~$\sqrt{\epsilon_2}$, the terms $\sqrt{\epsilon_2}\log(1-\sqrt{\epsilon_2})$, $\sqrt{\epsilon_2}\log(\frac{1 - \mu}{1-\sqrt{\epsilon_2}})$, $\sqrt{\epsilon_2}\log(\frac{\mu}{\sqrt{\epsilon_2}})$ become negligible compared to $\sqrt{\epsilon_2}\log(\sqrt{\epsilon_2})$, which is larger in magnitude.
Therefore, under the inverse consistency constraint, for any pixel $p$:
\begin{equation}\label{eq:entropy_gradient_upper}
H_B(p) \leq -2[\mu \log(\mu) + (1-\mu) \log(1-\mu)]+ \sqrt{\epsilon_2} \log(\sqrt{\epsilon_2}),
\end{equation}
where $\mu = D_{\text{fg}}(h)$.
\hfill \(\square\)

\noindent \textbf{Theorem 1 (Uncertainty Bound Under Constraints)} Under the following constraints: Consistency constraint for Architecture A: \( ||D_\text{fg1}(h) - D_\text{fg2}(h)||^2 \leq \epsilon_1 \) and  inverse consistency constraint for Architecture B: \( ||D_\text{fg}(h) + D_\text{bg}(h) - 1||^2 \leq \epsilon_2 \). There exists a constant \( C > 0 \) such that for sufficiently small \( \epsilon_1, \epsilon_2 > 0 \), we have:
\begin{equation}
H_B(p) \leq H_A(p) - C.
\end{equation}

\noindent \textbf{Proof:} From Lemma 1, the uncertainty in Architecture A is bounded  by Eq.~\eqref{eq:Lower_A}. From Lemma 2, the uncertainty in Architecture B is bounded  by Eq.~\eqref{eq:upper_b}. Assuming the shared encoder \( \mathcal{E} \) produces similar latent representations for both architectures, we have \( D_{\text{fg1}}(h) \approx D_{\text{fg}}(h) \), so \( \mu \) is comparable in both bounds. The difference between the upper bound of \( H_B(p) \) and the lower bound of \( H_A(p) \) is:
\begin{equation}
\begin{split}
H_B(p) - H_A(p) &\leq \sqrt{\epsilon_2} \log(\sqrt{\epsilon_2})+\sqrt{\epsilon_1} \log(\sqrt{\epsilon_1}).
\end{split}
\end{equation}
For small $\epsilon > 0$, the function $f(\epsilon) = \sqrt{\epsilon} \log(\sqrt{\epsilon})$ is negative and approaches 0. Its derivative is negative for small $\epsilon$, indicating $f(\epsilon)$ is decreasing.
Thus, for sufficiently small $\epsilon_1, \epsilon_2$, the sum
is bounded away from zero by a negative constant $-C$. Therefore:
\begin{equation}
H_B(p) \leq H_A(p) - C.
\end{equation}
\hfill \(\square\)
 
Through rigorous mathematical derivation, we have proven that Architecture B (foreground-background decoder) provides lower prediction uncertainty compared to Architecture A (dual foreground decoders). Additionally, the visualization results in Fig. 13 of the main paper further demonstrate that Architecture B effectively enhances the predictive confidence of the foreground model, as evidenced by distinct prediction boundaries.

Furthermore, in Theorem 2, under Architecture B, we elucidate the intrinsic mechanism by which background prediction reduces foreground prediction uncertainty.

\noindent \textbf{Notations:} We define the notation $\mathcal{L}_{\text{task}}$ as the task-specific loss function and $\mathcal{L}_\text{total}$ as the total loss function, which incorporates both the task loss and the inverse consistency loss. $q = D_{\text{bg}}(h)$ denotes background prediction. Consider the gradient of the total loss function with respect to the foreground prediction $\mu$:
\begin{equation}\label{eq:total_gradient} 
\begin{split}
 \frac{\partial \mathcal{L}_\text{total}}{\partial \mu} &= \frac{\partial \mathcal{L}_\text{task}}{\partial \mu} + \frac{\partial}{\partial \mu} (\mu + q - 1)^2\\
&=\frac{\partial \mathcal{L}_{\text{task}}}{\partial \mu} + 2(\mu + q - 1).
\end{split}
\end{equation}

The derivative of the foreground entropy is given by:
\begin{equation}\label{eq:entropy_gradient}
\begin{split}
\frac{\partial H(\mu)}{\partial \mu} 
&= -\frac{\partial}{\partial \mu} \left( \mu \log(\mu) \right) - \frac{\partial}{\partial \mu} \left( (1-\mu) \log(1-\mu) \right) \\
&= -\log(\mu) - 1 + \log(1-\mu) + 1 \\
&= \log\left(\frac{1-\mu}{\mu}\right)
\end{split}
\end{equation}
In gradient descent, the parameter update is:
\begin{equation}
\mu_{\text{new}} = \mu - \alpha \frac{\partial \mathcal{L}_{\text{total}}}{\partial \mu}
\end{equation}
where $\alpha > 0$ is the learning rate.

\noindent \textbf{Theorem 2 (Foreground Entropy Minimization)} When the background prediction significantly deviates from the median value, if the following conditions are satisfied:
\begin{equation}\label{eq:key_condition} 
\text{sign}\left(\frac{\partial \mathcal{L}_{\text{task}}}{\partial \mu}\right) = \text{sign}(\mu + q - 1), 
\end{equation}
and
\begin{equation}\label{eq:bg_condition} 
|q - 0.5| > |\mu - 0.5|,
\end{equation}
then the gradient update will reduce the entropy of the foreground prediction.

\noindent \textbf{Proof:} 
For the entropy to decrease after the update, we need:
\begin{equation}
H(\mu_{\text{new}}) < H(\mu)
\end{equation}

This occurs when $\mu_{\text{new}}$ moves away from 0.5 (since entropy is maximized at $\mu = 0.5$).
For $\mu < 0.5$, we need $\frac{\partial \mathcal{L}_{\text{task}}}{\partial \mu} > 0$ to make $\mu_{\text{new}} < \mu$.
For $\mu > 0.5$, we need $\frac{\partial \mathcal{L}_{\text{task}}}{\partial \mu} < 0$ to make $\mu_{\text{new}} > \mu$.

Let's analyze the two cases:

\noindent \textbf{Case 1: $\mu < 0.5$.} 
By invoking the inverse consistency constraint, we deduce that $q > 0.5$. Consequently, the inequality $|q - 0.5| > |\mu - 0.5|$ can be reformulated as $q - 0.5 > 0.5 - \mu$. 
This leads to the conclusion that $\mu + q > 1$. 
As a result, $\text{sign}(\mu + q - 1) > 0$, which implies that $\frac{\partial \mathcal{L}_{\text{task}}}{\partial \mu} > 0$. 
Given that $\mu < 0.5$, the positivity of $\frac{\partial \mathcal{L}_{\text{task}}}{\partial \mu}$ indicates that $\mu_{\text{new}} < \mu$, thereby leading to a reduction in entropy.

\noindent \textbf{Case 2: $\mu > 0.5$.} 
From the inverse consistency constraint, it follows that $q < 0.5$. Hence, the condition $|q - 0.5| > |\mu - 0.5|$ can be expressed as $0.5 - q > \mu - 0.5$. 
This implies that $\mu + q < 1$. 
Therefore, $\text{sign}(\mu + q - 1) < 0$, which signifies that $\frac{\partial \mathcal{L}_{\text{task}}}{\partial \mu} < 0$. 
Given that $\mu > 0.5$, the negativity of $\frac{\partial \mathcal{L}_{\text{task}}}{\partial \mu}$ suggests that $\mu_{\text{new}} > \mu$, which also results in a reduction in entropy.

In both cases, when the conditions of the theorem are satisfied, the gradient update moves the foreground prediction $\mu$ further away from 0.5, thereby reducing its entropy. \hfill \(\square\)

Therefore, when the background prediction significantly deviates from the median value (\ie, $|q - 0.5| > |\mu - 0.5|$) and $\text{sign}\left(\frac{\partial \mathcal{L}_{\text{task}}}{\partial \mu}\right) = \text{sign}(\mu + q - 1)$, the gradient update will reduce the entropy of the foreground prediction. 

Similarly, when the foreground decoder satisfies equivalent conditions, it also reduces the entropy of background predictions. This finding further elucidates the interrelationship between the predictions of cross-view models. Specifically, when the gradient directions of the task loss and inverse consistency loss satisfy the cosine similarity condition, \textit{low-uncertainty feature representations reduce the prediction uncertainty of high-uncertainty predictions through back-propagation within the cross-view models framework.}}
\end{document}